\documentclass[11pt]{article} 
\usepackage{hyperref}
\usepackage{url}
\usepackage{amssymb,amsfonts,amsmath,amsthm,amscd,dsfont,mathrsfs}
\usepackage{graphicx,float,psfrag,epsfig}
\usepackage{wrapfig}
\usepackage{color}

\newtheorem{propo}{Proposition}[section]
\newtheorem{lemma}[propo]{Lemma}

\newtheorem{coro}[propo]{Corollary}
\newtheorem{thm}{Theorem}
\newtheorem{rem}[propo]{Remark}

\makeatletter
\newsavebox{\@brx}
\newcommand{\llangle}[1][]{\savebox{\@brx}{\(\m@th{#1\langle}\)}%
  \mathopen{\copy\@brx\kern-0.5\wd\@brx\usebox{\@brx}}}
\newcommand{\rrangle}[1][]{\savebox{\@brx}{\(\m@th{#1\rangle}\)}%
  \mathclose{\copy\@brx\kern-0.5\wd\@brx\usebox{\@brx}}}
\makeatother

\def\<{\langle}
\def\>{\rangle}
\def\ones{\mathds 1}
\def\id{\mathbb I}
\newcommand{\prob}[1]{{ \mathbb{P}\left\{ #1 \right\} }}
\newcommand{\expect}[1]{\mathbb{E}\left[ #1 \right]}
\def\E{\mathbb E}
\def\reals{\mathbb R}
\def\Z{\mathbb Z}
\def\ind{\mathbb{I}}
\def\diag{{\rm diag}}

\newcommand{\vertiii}[1]{{\left\vert\kern-0.25ex\left\vert\kern-0.25ex\left\vert #1 
    \right\vert\kern-0.25ex\right\vert\kern-0.25ex\right\vert}}

\newcommand{\triplenorm}[1]{{\vert\kern-0.25ex \vert\kern-0.25ex \vert #1 
    \vert\kern-0.25ex \vert\kern-0.25ex \vert}}

\def\top{T}
    
\newcommand{\bb}{{\alpha}}
\newcommand{\hTheta}{\widehat{\Theta}}
\newcommand{\hL}{\widehat{L}}
\newcommand{\tell}{{\tilde \ell}}
\newcommand{\cL}{{\cal L}}
\newcommand{\cT}{{\cal T}}
\newcommand{\bB}{{\mathbb B}}
\newcommand{\calL}{{\cal L}}
\newcommand{\calA}{{\cal A}}
\newcommand{\calP}{{\cal P}}
\newcommand{\calB}{{\cal B}}
\newcommand{\calS}{{\cal S}}
\newcommand{\tV}{\widetilde{V}}
\newcommand{\tk}{\tilde{k}}
\newcommand{\tH}{\tilde{H}}
\newcommand{\tmu}{\tilde{\mu}}

\newcommand{\fnorm}[1]{\vertiii{#1}_{\rm F}}
\newcommand{\nucnorm}[1]{\vertiii{#1}_{\rm nuc}}

\newcommand{\lnorm}[2]{\vertiii{#1}_{{#2}}}
\newcommand{\Iprod}[2]{\llangle #1, #2 \rrangle}
\newcommand{\indc}[1]{\mathbb{I}\left( {#1 } \right) }

\title{  Collaboratively Learning Preferences from Ordinal Data }

\author{
{Sewoong Oh\thanks{Department of Industrial and Enterprise Systems Engineering, University of Illinois at Urbana-Champaign, e-mail: \texttt{swoh@illinois.edu}},  
Kiran K. Thekumparampil\thanks{Department of Electrical and Computer Engineering, University of Illinois at Urbana-Champaign, e-mail: \texttt{thekump2@illinois.edu}},
and Jiaming Xu\thanks{Statistics Department, University of Pennsylvania, e-mail: \texttt{jxu18@illinois.edu}} 
}
}

\begin{document}

\maketitle

\begin{abstract}
In applications such as recommendation systems and revenue management, 
it is important to predict preferences on items that have not been seen by a user or 
predict outcomes of comparisons among those that have never been compared. 
A popular discrete choice model of multinomial logit  model 
captures  the structure of the hidden preferences  with a low-rank matrix. 
In order to predict the preferences, we want to learn the underlying model from 
noisy observations of the low-rank matrix, collected as 
revealed preferences in various forms of ordinal data. 
A natural approach to learn such a   model 
 is to solve a convex relaxation of nuclear norm minimization. 
We present the convex relaxation approach   in two contexts of interest: collaborative ranking and 
bundled choice modeling. 
In both cases, we show that the convex relaxation is minimax optimal. 
We 
prove  
an upper bound on the resulting error with finite samples, and  
provide a matching information-theoretic lower bound. 
\end{abstract}

%
%
%
%
%

\section{Introduction} 
\label{sec:intro} 




In many applications such as recommendation systems and revenue management, 
it is important to predict preferences on items that have not been seen by a user or 
predict outcomes of comparisons among those that have never been compared. 
Predicting such hidden preferences 
 would be hopeless without further assumptions on the structure of the preference. 
Motivated by the success of matrix factorization models 
on collaborative filtering applications, 
we model hidden preferences with 
 low-rank matrices to collaboratively learn preference matrices from ordinal data. 
In this paper, we consider the following two concrete scenarios: 
\vspace{-0.3cm}
\begin{itemize} 
	\item {\em Collaborative ranking.} 
		Consider an online market that 
		collects each user's preference as a ranking over a subset of items that are `seen' by the user. 
		Such data can be obtained by directly asking to compare some items, or 
		by indirectly tracking online activities on 
		which items are viewed, how much time is spent on the page, or 
		how the user rated the items. 		
		In order to make personalized recommendations, 
		we want $(a)$ a model that captures how users who preferred similar items are also likely to 
		have similar preferences on unseen items; and $(b)$ 
		 to predict which items the user might prefer, by learning such models from ordinal data. 
	\item {\em Bundled choice modeling.} 
		Discrete choice models describe how a user makes decisions on what to purchase. 
		Typical choice models assume the willingness to buy an item is independent of what else the user bought. 
		In many cases, however, we make `bundled' purchases: 
		we buy particular ingredients together for one recipe or  we buy two connecting flights. 
		One choice (the first flight) has a significant impact on the other (the connecting flight). 	
		In order to optimize the assortment (which flight schedules to offer) for maximum expected revenue, 
		it is crucial to accurately predict the willingness of the consumers to purchase, 
		based on past history.   
		We consider a case where there are two types of products (e.g. jeans and shirts), 
		and want 
		$(a)$ a  model that captures such interacting preferences for pairs of items, 
		one from each category; 		 
		and $(b)$ to predict the consumer's choice probabilities on pairs of items, by learning such models from past purchase history. 
\end{itemize} 

We use a discrete choice model known as MultiNomial Logit (MNL) model (described in Section \ref{sec:MNL}) 
to represent the preferences. 
In collaborative ranking context, 
MNL uses a low-rank matrix to represent 
the hidden preferences of the users. 
Each row corresponds to a user's preference over all the items, and 
when presented with a subset of items the user provides a ranking over those items, 
which is a noisy version of the hidden true preference. 
the low-rank assumption naturally captures the similarities among users and items, by representing each on a low-dimensional space. 
In bundled choice modeling context, the low-rank matrix now represents 
how pairs of items are matched. Each row corresponds to an item from the first category 
and each column corresponds to an item from the second category. 
An entry in the matrix represents how much the pair is preferred by a randomly chosen user from a pool of users. 
Notice that in this case we do not model individual preferences, but the preference of the whole population. 
The purchase history of the population is the record of which pair was chosen among  a subsets of items that were presented, 
which is again a noisy version of the hidden true preference. 
The low-rank assumption captures the 
similarities and dis-similarities among the items in the same category and 
the interactions across categories. 

{\bf Contribution.}
A natural approach to learn such a low-rank model, 
from noisy observations, is to solve a convex relaxation of nuclear norm minimization (described in Section \ref{sec:opt}). 
We present such an approach for learning the MNL model from ordinal data, in two contexts: collaborative ranking and 
bundled choice modeling. In both cases, we analyze the sample complexity of the algorithm, and provide 
an upper bound on the resulting error with finite samples. 
We prove minimax-optimality of our approach by 
providing a matching information-theoretic lower bound (up to a poly-logarithmic factor). 
Technically, we utilize the Random Utility Model (RUM) interpretation (outlined in Section \ref{sec:MNL}) of the MNL model to 
prove both the upper bound and the fundamental limit, which could be of interest to analyzing more general class of RUMs.   

{\bf Related work.}
In the context of collaborative ranking, 
MNL models have been proposed to model partial rankings from a pool of users. 
Existing work is limited to the case when each user provides pair-wise comparisons \cite{LN14,PNZSD15}. 
\cite{PNZSD15} proposes solving a convex relaxation of maximizing the likelihood 
over matrices with bounded nuclear norm. 
It is shown that this approach achieves statistically optimal generalization error rate. 
Our analysis techniques are inspired by 
\cite{LN14}, which proposed the convex relaxation similar to ours, but when the users provide only pair-wise comparisons.  
For pairwise comparisons, our main result in Theorem \ref{thm:bundle_ub} matches those of \cite{LN14}, 
but our result is more general in the sense that we analyze more general sampling models beyond pairwise comparisons. 
In general, ``collaborative ranking'' has been used typically to refer to the problem of learning personal rankings when 
the data is ratings on items (as opposed to ordinal data). 
Matrix factorization approaches have been widely applied in practice \cite{YJJJ13,RFGS09}, 
but no theoretical guarantees are known. 




The remainder of the paper is organized as follows. 
In Section \ref{sec:model}, we present the MNL model and 
propose a convex relaxation for learning the model, in the context of collaborative ranking.  
We provide theoretical guarantees for collaborative ranking  in Section \ref{sec:kwise}. 
In Section \ref{sec:bundle}, we present the problem statement for bundled choice modeling, and 
analyze a similar convex relaxation approach. 

{\bf Notations.} 
We use $\fnorm{A}$
and  $\lnorm{A}{\infty}$
to denote  the 
Frobenius norm and the $\ell_\infty$ norm, 
$\nucnorm{A}=\sum_{i}\sigma_i(A)$ to denote the nuclear norm 
where $\sigma_i(A)$ denote the $i$-th singular value, 
and $\lnorm{A}{2}=\sigma_1(A)$ for the spectral norm.
We use $\llangle u,v \rrangle = \sum_i u_iv_i$ 
and $\|u\|$ to denote the inner product and the Euclidean norm. 
All ones vector is denoted by $\ones$ and 
$\ind{(A)}$ is the indicator function of the event $A$.
The set of the fist $N$ integers are denoted by $[N]=\{1,\ldots,N\}$.

\section{Model and Algorithm}
\label{sec:model}
In this section, we present a discrete choice modeling for collaborative ranking, and 
propose an inference algorithm for learning the model from ordinal data. 

\subsection{MultiNomial Logit (MNL) model for comparative judgment}
\label{sec:MNL}

In collaborative ranking, we want to model how people who have similar preferences on a subset of items 
are likely to have similar tastes on other items as well. 
When users provide  ratings, as in collaborative filtering applications, 
matrix factorization models are widely used 
since the low-rank structure captures the similarities between users. 
When users provide ordered preferences, 
we use a discrete choice  model known as MultiNomial Logit (MNL) model that has 
a similar low-rank structure that captures the similarities between users and items. 

Let $\Theta^*$ be the $d_1\times d_2$ dimensional matrix capturing 
the preference of $d_1$users on $d_2$ items, where the 
rows and columns correspond to users and items, respectively. 
Typically, $\Theta^*$ is assumed to be low-rank, having a rank $r$ that is much smaller than the dimensions. 
However, in the following we allow a more general setting where $\Theta^*$ might be only approximately low rank.  
When a user $i$ is 
presented with a set of alternatives $S_i\subseteq [d_2]$, 
she reveals her preferences 
as  a  ranked list over those items. 
To simplify the notations we assume all users compare the same number $k$ of items, 
but the analysis naturally generalizes to the case when the size might differ from  a user to a user. 
Let $v_{i,\ell} \in S_i$ denote the (random) $\ell$-th best choice of user $i$. 
Each user gives a ranking,  independent of other users' rankings, from 
\begin{align}
	 \prob{v_{i,1},\ldots,v_{i,k}} & = \, \prod_{\ell = 1}^k \frac{e^{\Theta^*_{i,v_{i,\ell}}}}{\sum_{j\in S_{i,\ell}} e^{\Theta^*_{i,j}} } 
	\;, \label{eq:defkwiseMNL}
\end{align}
where 
with $S_{i,\ell} \equiv S_i \setminus \{ v_{i,1}, \ldots, v_{i,\ell-1} \}$. 
For a single user $i$, the $i$-th row of $\Theta^*$ 
represents the underlying preference vector of 
the user, and the more preferred items are more likely to be ranked higher.   
The probabilistic nature of the model  captures the noise in the users revealed preferences.

The random utility model (RUM), pioneered by  \cite{Thu27,Mar60,Luce59},  describes the 
choices of users as manifestations of the underlying utilities. 
The MNL models is a special case of RUM where 
each decision maker and each alternative 
are represented by a $r$-dimensional feature vectors  
$u_i $ and $v_j$ respectively, such that $\Theta^*_{ij}=\llangle u_i, v_j \rrangle$, resulting in a low-rank matrix.  
When presented with a set of alternatives $S_i$, 
the decision maker $i$ ranks the alternatives according to their random utility drawn from 
\begin{eqnarray}
	U_{ij} &=& \llangle u_i,v_j \rrangle + \xi_{ij}\;, \label{eq:defkwiseRUM}
\end{eqnarray}
for item $j$, where $\xi_{ij}$ follow 
the standard Gumbel distribution. 
Intuitively, this provides a justification for the MNL model 
as modeling the decision makers as rational being, seeking to maximize utility.
Technically,  this RUM interpretation   plays a crucial role in our analysis, in proving restricted strong convexity in Appendix \ref{sec:kwise_hessian3_proof} 
and also in proving fundamental limit in Appendix \ref{sec:kwise_lb_proof}.

There are a few cases where the Maximum Likelihood (ML) estimation for RUM is tractable. 
One notable example is the Plackett-Luce (PL) model, which is a special case of the MNL model where $\Theta^*$ is rank-one and all users have the same features.  
PL model has been widely applied in econometrics \cite{McF73}, 
analyzing elections \cite{GM09}, and machine learning \cite{Liu09}.  
Efficient inference algorithms has been proposed \cite{Hun04,GS09,CD12}, 
and the sample complexity has been analyzed for the MLE \cite{HOX14} and 
for the Rank Centrality \cite{NOS12}. 
Although PL is quite restrictive, in the sense that it assumes all users share the same features,  
little is known about inference in RUMs beyond PL. 
Recently, to overcome such a restriction, 
mixed PL models 
have been studied, where 
$\Theta^*$ is rank-$r$ but there are only $r$ classes of users and 
all users in the same class have the same features. 
Efficient inference algorithms with provable guarantees have been proposed 
by applying recent advances in tensor decomposition methods 
\cite{OS14,DIS14},  
directly clustering the users 
\cite{AOSV14,WXSMLH15}, or using sampling methods \cite{ADLP13}.
However, this mixture PL is still restrictive, and 
both clustering and tensor based approaches rely heavily on the fact that the distribution is a ``mixture'' and require additional incoherence assumptions on $\Theta^*$.
For more general models, 
efficient inference algorithms have been proposed \cite{SPX12} 
but no performance guarantee is known for finite samples. 
Although the MLE for 
the general MNL model in \eqref{eq:defkwiseMNL} is intractable, 
we  
provide a polynomial-time inference algorithm with provable guarantees.

\subsection{Nuclear norm minimization}
\label{sec:opt}

Assuming $\Theta^*$ is well approximated by a low-rank matrix, 
we estimate $\Theta^*$ by solving the following convex relaxation 
given the observed preference in the form of ranked lists $\{(v_{i,1},\ldots,v_{i,k} )\}_{i\in[d_1]}$. 
\begin{align}
\widehat{\Theta} \,\in\, \arg \min_{\Theta \in \Omega} \,\cL(\Theta) \,+\, \lambda \nucnorm{\Theta},
\label{eq:kwiseopt}
\end{align} 
where the (negative) log likelihood function according to \eqref{eq:defkwiseMNL} is 
\begin{align}
	\cL(\Theta)\,=\, -\frac{1}{k\, d_1} \sum_{i=1}^{d_1}  \sum_{\ell=1}^{k} \left( \llangle \Theta, e_{i} e_{v_{i,\ell} }^\top    \rrangle - \log \left( \sum_{j \in S_{i,\ell} } \exp\left( \llangle \Theta, e_{i} e_j^\top \rrangle \right) \right) \right) \;,
	\label{eq:defkwiseL}
\end{align}
with $S_i=\{v_{i,1},\ldots,v_{i,k}\}$ and  $S_{i,\ell} \equiv S_i \setminus \{ v_{i,1}, \ldots, v_{i,\ell-1} \}$, 
and appropriately chosen set $\Omega$ defined in \eqref{eq:defkwiseomega}.
Since nuclear norm is a tight convex surrogate for the rank, 
the above optimization searches for a low-rank solution that maximizes the likelihood. 
Nuclear norm minimization has been widely used in 
rank minimization problems \cite{RFP10}, but  
provable guarantees typically exists only for quadratic loss function $\calL(\Theta)$ 
\cite{CR09,NW11}. Our analysis extends such analysis techniques to identify 
the conditions under which 
restricted strong convexity is satisfied for a convex loss function that is not quadratic. 


\section{Collaborative ranking from $k$-wise comparisons} 
\label{sec:kwise}

We first provide background on the MNL model, and then present main results on the performance guarantees.  
Notice that the distribution \eqref{eq:defkwiseMNL} is independent of shifting each row of $\Theta^*$ by a constant. Hence, there is an equivalent class of $\Theta^*$ that gives the same  distributions for the ranked lists:  
\begin{eqnarray}
	[\Theta^*] = \{A \in \reals^{d_1\times d_2} \;|\; A = \Theta^* + u\,\ones^T \text{ for some }u\in \reals^{d_1}\}\;. 
\end{eqnarray} 
Since we can only estimate $\Theta^*$ up to this equivalent class, we search for the one 
whose rows sum to zero, i.e. $\sum_{j\in[d_2]} \Theta^*_{i,j}=0$ for all $i\in[d_1]$.
Let  
$\alpha \equiv \max_{i, j_1,j_2}  |\Theta_{ij_1}^*- \Theta^*_{ij_2}| $ 
denote the dynamic range of 
the underlying $\Theta^*$, such that when $k$ items are compared, we always have 
\begin{eqnarray}
 \frac1k e^{-\bb} \;\leq\; \prob{v_{i,1} = j} \;\leq\; \frac1k  e^\bb \;,
 \end{eqnarray}
for all $j\in S_i$, all $S_i\subseteq [d_2]$ satisfying $|S_i|=k$ and all $i\in[d_1]$.  
We do not make any assumptions on $\bb$ other than that 
$\bb=O(1)$ with respect to $d_1$ and $d_2$. 
The purpose of defining the dynamic range in this way is that we seek to characterize how 
the error scales with $\bb$. 
Given this definition, we solve the optimization in \eqref{eq:kwiseopt} over 
 \begin{eqnarray}
 	\Omega _\bb = \Big\{ A \in \reals^{d_1 \times d_2} \,\big|\, \lnorm{A}{\infty} \le \bb, \text{ and }\forall i\in[d_1]\text{ we have }\sum_{j\in[d_2]} A_{ij}=0 \Big\} \;.\label{eq:defkwiseomega}
\end{eqnarray}
While in practice we do not require the $\ell_\infty$ norm constraint,  
we need it for the analysis. 
For a related problem of matrix completion, where the loss $\calL(\theta)$ is quadratic, 
either a similar condition on $\ell_\infty$ norm is required or 
a different condition on incoherence is required.

\subsection{Performance guarantee} 

We provide an upper bound on the resulting error of our convex relaxation, 
when a {\em multi-set} of items $S_i$ presented to user $i$ is drawn uniformly at random 
with replacement. 
Precisely, for a given $k$, $S_i = \{ j_{i,1},\ldots, j_{i,k}\}$ where 
$j_{i,\ell}$'s are independently drawn uniformly at random over the $d_2$ items. 
Further, if an item is sampled more than once, i.e. if  there exists 
$j_{i,\ell_1}=j_{i,\ell_2}$ for some $i$ and $\ell_1\neq \ell_2$, then we assume that the user 
treats these two items as if they are two distinct items with the same MNL weights 
$\Theta_{i,j_{i,\ell_1}}^*=\Theta_{i,j_{i,\ell_2}}^*$.The resulting preference is therefore always over $k$ items 
(with possibly multiple copies of  the same item), and distributed according to \eqref{eq:defkwiseMNL}. 
For example, if $k=3$, it is possible to have $S_i=\{j_{i,1}=1,j_{i,2}=1,j_{i,3}=2\}$, in which case the resulting 
ranking can be $ (v_{i,1} = j_{i,1},v_{i,2}=j_{i,3}, v_{i,3}=j_{i,2} )$ with probability $ (e^{\Theta^*_{i,1}})/(2\,e^{\Theta^*_{i,1}}+e^{\Theta^*_{i,2}}) \times (e^{\Theta^*_{i,2}})/(e^{\Theta^*_{i,1}}+e^{\Theta^*_{i,2}}) $. 
Such sampling with replacement is necessary for the analysis, where 
we require independence in the choice of the items in $S_i$ 
in order to apply the symmetrization technique (e.g. \cite{BLN13}) to bound the expectation of the deviation (cf. Appendix  \ref{sec:kwise_hessian3_proof}).  Similar sampling assumptions have been made in existing analyses on learning 
low-rank models from noisy observations, e.g. \cite{NW11}. 
Let $d \equiv (d_1+d_2)/2$, and  let $\sigma_j(\Theta^*)$ denote the $j$-th singular value of the matrix $\Theta^*$. Define  
\begin{eqnarray*}
	\lambda_0 &\equiv& e^{2\bb} \sqrt{\frac{d_1\log d + d_2(\log d)^2}{k\,d_1^2 \,d_2}} \;.  
\end{eqnarray*} 

\begin{thm}
\label{thm:kwise_ub}
Under the described sampling model, assume $24  \, \leq k \leq \min\{ d_1^2, (d_1^2 + d_2^2)/(2d_1) \} \log d$, and
$\lambda \in [32 \lambda_0, c_0 \lambda_0]$
with  any constant $c_0=O(1)$ larger than 32. Then, solving the optimization \eqref{eq:kwiseopt} achieves 
\begin{eqnarray}
		\frac{1}{d_1d_2} \fnorm{\hTheta-\Theta^\ast}^2  \; \leq \; 288\sqrt{2}  \, e^{4\bb}  c_0 \lambda_0 \sqrt{r} \,  \fnorm{ \hTheta-\Theta^\ast } + 288 e^{4\bb} c_0 \lambda_0 \sum_{j=r+1}^{\min\{d_1,d_2\}} \sigma_j(\Theta^*) \;, 
	\label{eq:kwise_ub}
\end{eqnarray}
for any $r\in \{1,\ldots, \min\{d_1,d_2\} \}$ 
with probability at least $1-2d^{-3}$ where $d=(d_1+d_2)/2$.
\end{thm}

A proof   is provided in Appendix \ref{sec:kwise_ub_proof}. 
The above bound shows a natural splitting of the error into two terms, one corresponding to the {\em estimation error} 
for the rank-$r$ component and the second one corresponding to the {\em approximation error} for how well one can approximate $\Theta^*$ with a rank-$r$ matrix. 
This bound holds for all values of $r$ and one could potentially optimize over $r$. 
We show such results in the following corollaries.

\begin{coro}[{\bf Exact low-rank matrices}]
	Suppose $\Theta^*$ has rank at most $r$. 
	Under the hypotheses of Theorem \ref{thm:kwise_ub}, 
	solving the optimization \eqref{eq:kwiseopt} with the choice of the regularization parameter 
	$\lambda\in[32\lambda_0,c_0\lambda_0]$ achieves 	with probability at least $1-2d^{-3}$, 
	\begin{eqnarray}
	\frac{1}{\sqrt{d_1d_2}} \fnorm{\widehat{\Theta}-\Theta^\ast} \,\leq\, 
	288 \sqrt{2}   e^{6 \bb}  c_0 \sqrt{\frac{r(d_1\log d + d_2(\log d)^2) }{k\,d_1}} \;.  
	\label{eq:kwise_lowrank}
	\end{eqnarray}
	\label{cor:kwise_lowrank}
\end{coro}

The number of entries is $d_1d_2$ and we rescale the Frobenius norm error appropriately by $1/\sqrt{d_1d_2}$. 
When $\Theta^*$ is a rank-$r$ matrix, then
the degrees of freedom  in representing $\Theta^*$ is
$r(d_1+d_2)-r^2 = O(r(d_1+d_2))$.
The above theorem shows that the total number of samples, which is
$(k \, d_1)$, needs to scale as $O(rd_1(\log d) + r d_2 (\log d)^2)$ 
in order to achieve an arbitrarily small error. This is only poly-logarithmic 
factor larger than the degrees of freedom. 
In Section \ref{sec:kwise_lb}, we provide a lower bound on the error directly, that matches the upper bound up to a logarithmic factor.

\begin{figure}
	\begin{center}
	\includegraphics[width=.45\textwidth]{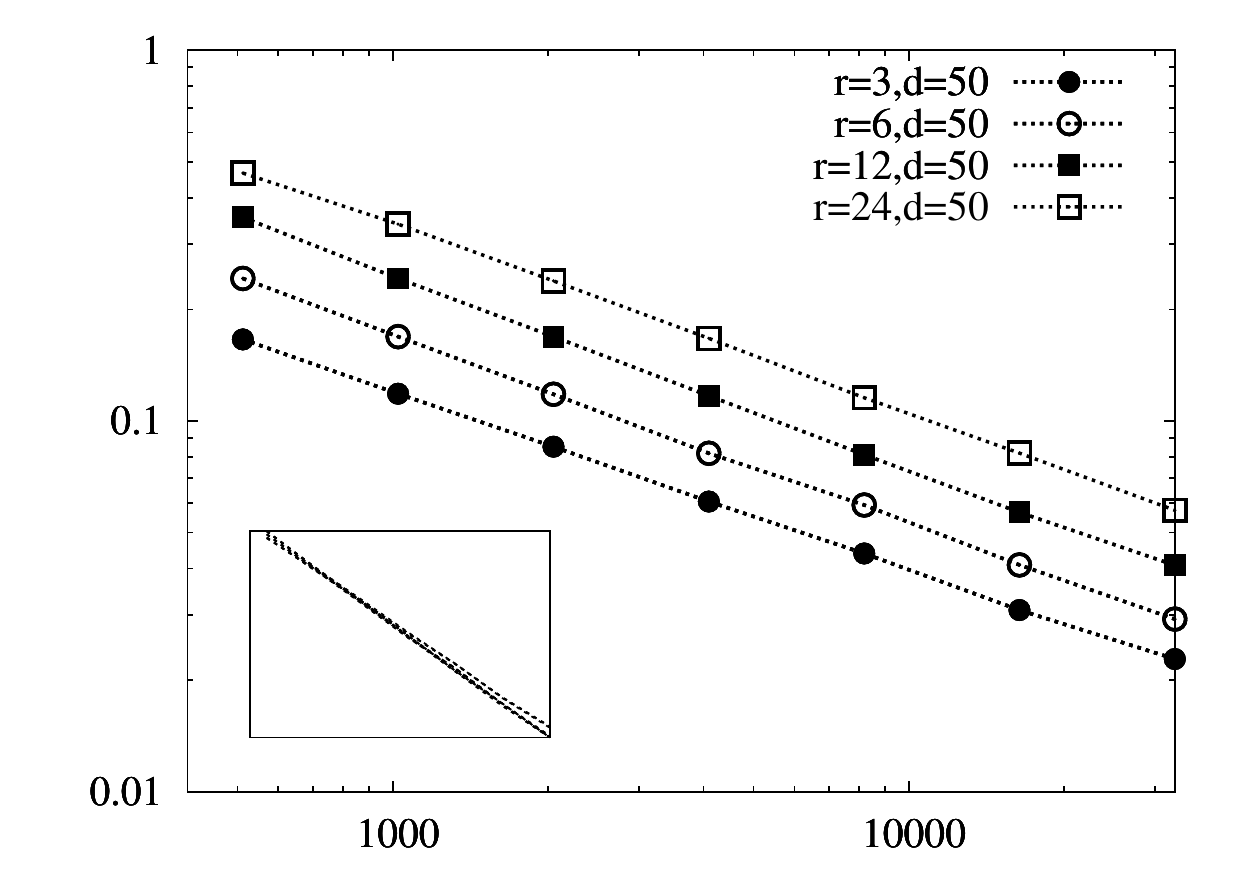}
	\put(-195,149){\small RMSE}
	\put(-125,-5){\small sample size $k$}
	\hspace{0.4cm}
	\includegraphics[width=.45\textwidth]{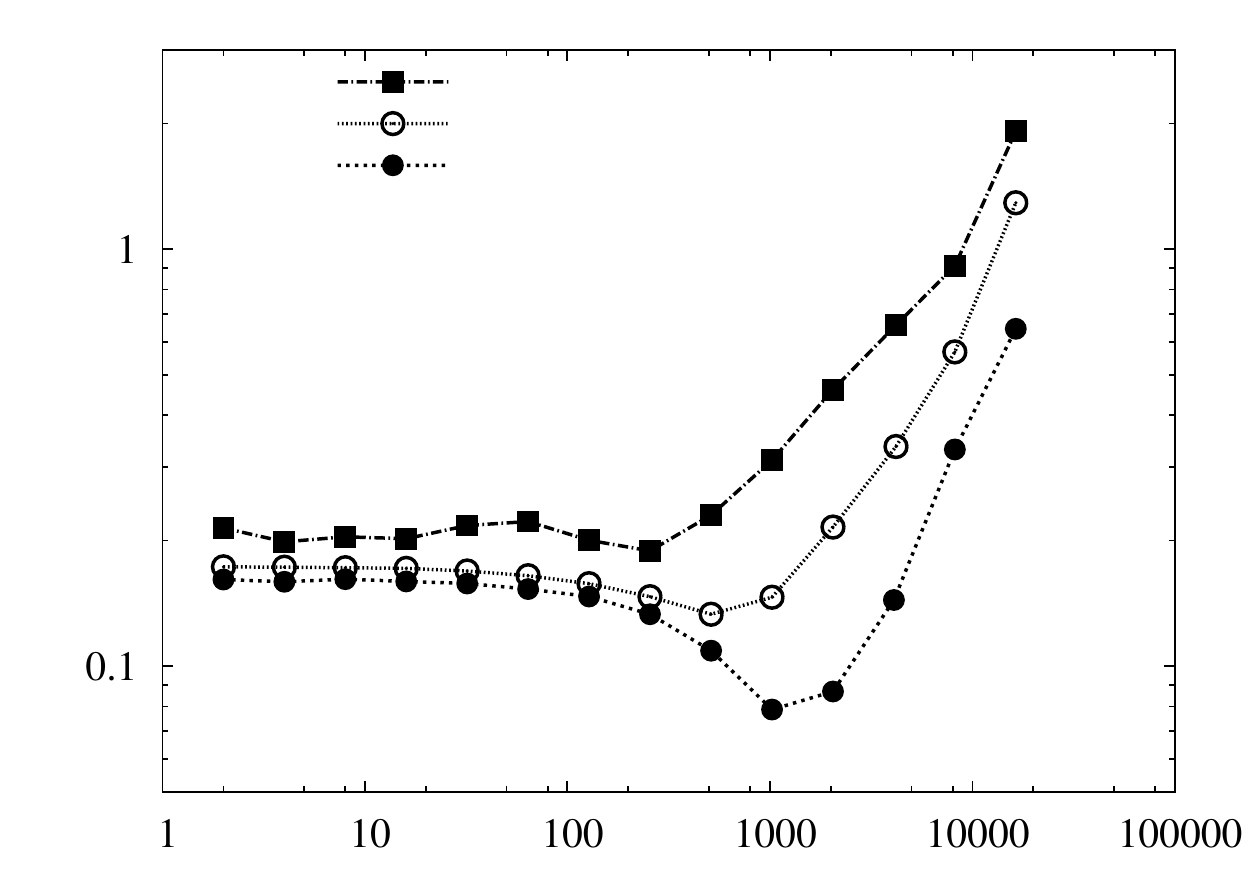}
	\put(-195,149){\small RMSE}
	\put(-129,-5){\small $\frac{\lambda}{\sqrt{(\log d) / (k d^2)}}$}
	\put(-134,133){\tiny $\bb=15$}
	\put(-134,126){\tiny $\bb=10$}
	\put(-134,119){\tiny $\bb=5$}
	\end{center}
	\caption{ The (rescaled) RMSE scales as $\sqrt{r(\log d)/k}$ as expected from Corollary  \ref{cor:kwise_lowrank} for fixed $d=50$ (left).  In the inset, the same data is plotted versus rescaled sample size $k/(r\log d)$. 
			The (rescaled) RMSE is stable for a broad range of $\lambda$ and $\bb$ for fixed $d=50$ and $r=3$ 
			(right).  }
	\label{fig:kwise}
\end{figure}
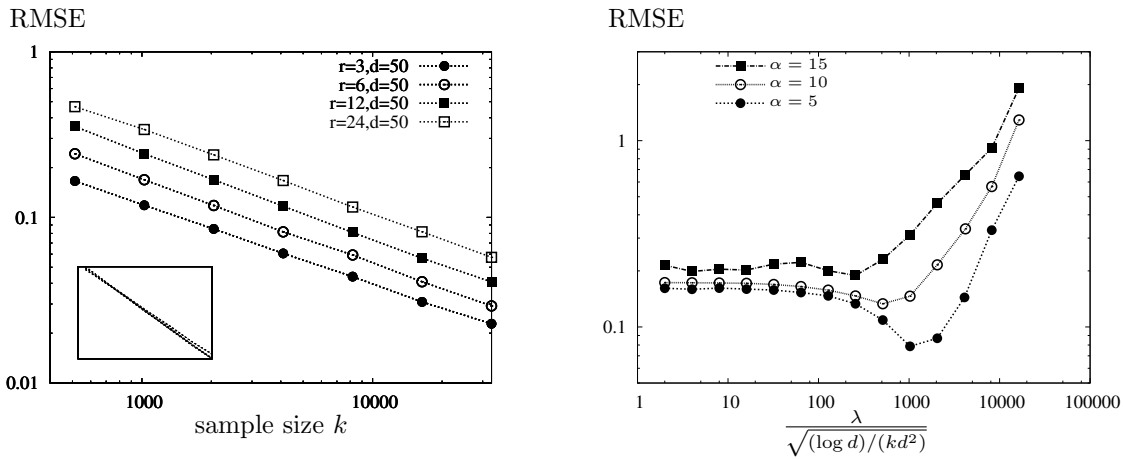

The dependence 
on the dynamic range $\bb$, however, is sub-optimal. 
It is expected that the error increases with $\bb$, since the $\Theta^*$ scales as $\bb$, but the exponential dependence in the bound 
seems to be a weakness of the analysis, as seen from numerical experiments in the right panel of Figure \ref{fig:kwise}. Although the error increase with $\bb$, numerical experiments suggests that it only increases at most linearly. 
However, tightening the scaling with respect to $\bb$ is a challenging problem, and 
such sub-optimal dependence is also present in 
existing literature for learning even  simpler models, such as the Bradley-Terry model \cite{NOS12} 
 or the Plackett-Luce model \cite{HOX14}, which are special cases of the MNL model studied in this paper.  
A practical issue in achieving the above rate is the choice of $\lambda$, since 
the dynamic range $\bb$ is not known in advance. 
Figure \ref{fig:kwise} illustrates that the error is not sensitive to the choice of $\lambda$ for a wide range. 

Another issue is that the underlying matrix might not be exactly low rank. 
It is more realistic to assume that it is approximately low rank. 
Following \cite{NW11} we formalize this notion with ``$\ell_q$-ball'' of matrices defined as 
\begin{eqnarray}
	\bB_q(\rho_q) &\equiv & \{ \Theta\in\reals^{d_1\times d_2} \,|\, \sum_{j\in[\min\{d_1,d_2\}]}^{} |\sigma_j(\Theta^*)|^q\leq \rho_q  \} \;. \label{eq:defBq}
\end{eqnarray}
When $q=0$, this is a set of rank-$\rho_0$ matrices. 
For $q\in(0,1]$, this is set of matrices whose singular values decay relatively fast. 
Optimizing the choice of $r$ in Theorem \ref{thm:kwise_ub}, we get the following result. 

\begin{coro}[{\bf Approximately low-rank matrices}]
	Suppose $\Theta^* \in \bB_q(\rho_q)$ for some $q\in(0,1]$ and $\rho_q>0$. 
	Under the hypotheses of Theorem \ref{thm:kwise_ub}, 
	solving the optimization \eqref{eq:kwiseopt} with the choice of the regularization parameter 
	$\lambda \in [32\lambda_0,c_0\lambda_0]$ achieves 
		with probability at least $1-2d^{-3}$, 
	\begin{eqnarray}
	\frac{1}{\sqrt{d_1d_2}} \fnorm{\widehat{\Theta}-\Theta^\ast} \,\leq\, 
	 \frac{2\sqrt{\rho_q}}{\sqrt{d_1d_2}}  \left(  288 \sqrt{2}  c_0 e^{6 \bb} \,\sqrt{\frac{d_1d_2(d_1\log d + d_2(\log d)^2) }{k\,d_1}} \right)^{\frac{2-q}{2}}\;. 
	\label{eq:kwise_appxlowrank}
	\end{eqnarray}
	\label{cor:kwise_appxlowrank}
\end{coro}

This is a strict generalization of Corollary \ref{cor:kwise_lowrank}. 
For $q=0$ and $\rho_0=r$, this recovers the exact low-rank estimation bound up to a factor of two. 
For approximate low-rank matrices in an $\ell_q$-ball, we lose in the error exponent, which reduces from one to $(2-q)/2$. A proof of this Corollary is provided in Appendix \ref{sec:kwise_cor_proof}. 

The left  panel of Figure \ref{fig:kwise} confirms the scaling of the error rate as predicted by Corollary \ref{cor:kwise_lowrank}. 
The lines merge to a single line when the sample size is rescaled appropriately. 
We make a choice of $\lambda= (1/2) \sqrt{ (\log d)/(kd^2)}$, 
This choice is independent of $\bb$ and is smaller than proposed in Theorem \ref{thm:kwise_ub}. 
We generate random rank-$r$ matrices of dimension $d\times d$, where 
$\Theta^*=UV^T$ with $U\in\reals^{d\times r}$ and $V\in\reals^{d\times r}$ entries generated i.i.d 
from uniform distribution over $[0,1]$. 
Then the row-mean is subtracted form each row, and then the whole matrix is scaled such that the largest entry is $\bb=5$.  
The root mean squared error (RMSE) is plotted where 
${\rm RMSE} = (1/d) \triplenorm{ \Theta^*-\hTheta}_{\rm F} $.   
We implement and solve the convex optimization \eqref{eq:kwiseopt} 
using proximal gradient descent method as analyzed in \cite{ANW10}. 
The right panel in Figure \ref{fig:kwise} illustrates that the actual  
error is insensitive to the choice of $\lambda$ for a broad range of 
$\lambda\in[\sqrt{ (\log d)/(kd^2)},2^{8}\sqrt{ (\log d)/(kd^2)}]$, after which it increases with $\lambda$.

\subsection{Information-theoretic lower bound for  low-rank matrices } 
\label{sec:kwise_lb} 

For a polynomial-time algorithm of convex relaxation, we gave in the previous section a bound on the  achievable error. We next compare this to the fundamental limit of this problem, 
by giving a lower bound on the achievable error  by any algorithm (efficient or not). 
A simple parameter counting argument indicates that 
it requires the number of samples to scale as the degrees of freedom i.e.,   
$k d_1 \propto r(d_1+d_2)$, 
to estimate a $d_1\times d_2$ dimensional matrix of rank $r$. 
We construct an appropriate packing over the set of low-rank matrices with 
bounded entries in $\Omega_\bb$ defined as \eqref{eq:defkwiseomega}, 
and  show that no algorithm can accurately estimate 
 the true matrix with high probability using the generalized Fano's inequality. 
This provides a constructive argument to lower bound the minimax error rate, which in turn establishes that the bounds in Theorem \ref{thm:kwise_ub} is sharp up to a logarithmic factor, and proves no other algorithm can significantly improve over the nuclear norm minimization. 

\begin{thm}
Suppose $\Theta^*$ has rank $r$. 
Under the described sampling model, for large enough $d_1$ and $d_2\geq d_1$, 
there is a universal numerical constant $c>0$ such that 
\begin{eqnarray}
	\inf_{\hTheta} \sup_{\Theta^*\in\Omega_\bb} \E\Big[ \frac{1}{\sqrt{d_1d_2}}\fnorm{\hTheta-\Theta^*} \Big] &\geq& c\,\min\left\{ \bb e^{-\bb} \sqrt{\frac{\,r\,d_2}{k\,d_1}}  \,,\,  \frac{\bb d_2}{\sqrt{d_1d_2 \log d}}\right\} \;,
\end{eqnarray}
where the infimum is taken over all measurable functions over the observed ranked lists  
$\{(v_{i,1},\ldots,v_{i,k})\}_{i\in[d_1]}$. 
\label{thm:kwise_lb}
\end{thm}
A proof of this theorem is provided in Appendix \ref{sec:kwise_lb_proof}. 
The term of primary interest in this bound is the first one, 
which shows the scaling of the (rescaled) minimax rate as $\sqrt{r(d_1+d_2)/(kd_1)}$ (when $d_2\geq d_1$), and matches the upper bound in \eqref{eq:kwise_ub}.
It is the dominant term in the bound whenever 
the number of samples is larger than the degrees of freedom by a logarithmic factor, i.e.,  
$kd_1 >  r(d_1+d_2) \log d$, ignoring the dependence on $\bb$.
This is a typical regime of interest, where the sample size is comparable to the latent dimension of the problem.
In this regime,  
Theorem \ref{thm:kwise_lb} establishes that the upper bound in 
Theorem \ref{thm:kwise_ub} is minimax-optimal up to a logarithmic factor in the dimension $d$.

\section{Choice modeling for bundled purchase history} 
\label{sec:bundle}

In this section, we use the MNL model to 
study another scenario of practical interest: choice modeling from bundled purchase history. 
In this setting, we assume that we have bundled purchase history data from $n$ users.  
Precisely, there are two categories of interest with $d_1$ and $d_2$ alternatives in each category respectively. For example, there are $d_1$ tooth pastes to choose from and $d_2$ tooth brushes to choose from. 
For the $i$-th user, a subset $S_i \subseteq [d_1]$ of alternatives from the first category is presented along with a subset $T_i\subseteq[d_2]$ of alternatives from the second category. 
We use $k_1$ and $k_2$ to denote the number of alternatives presented to a single user, i.e. $k_1=|S_i|$ and $k_2=|T_i|$, and we assume that the number of alternatives presented to each user is fixed, to simplify notations. 
Given these sets of alternatives, each user makes a `bundled' purchase 
and we use $(u_i,v_i)$ to denote the bundled pair of alternatives (e.g. a tooth brush and a tooth paste) purchased by the $i$-th user. 
Each user makes a choice of the best alternative, independent of other users's choices, according to the MNL model as 
\begin{eqnarray}
	\prob{(u_i,v_i)=(j_1,j_2)} &=& \frac{e^{\Theta^*_{j_1,j_2}}}{ \sum_{j'_1\in S_i,j'_2\in T_i} e^{\Theta^*_{j_1',j_2'}}}\;,
	\label{eq:defbundleMNL}
\end{eqnarray}
for all $j_1\in  S_i$ and $j_2 \in T_i$. 
The distribution \eqref{eq:defbundleMNL} is independent of shifting all the values of $\Theta^*$ by a constant. Hence, there is an equivalent class of $\Theta^*$ that gives the same distribution for the choices:
	$ [\Theta^*] \equiv \{A \in \reals^{d_1\times d_2}\,|\, A = \Theta^* + c \ones \ones^T \text{ for some }c\in\reals  \}\;.$ 
Since we can only estimate $\Theta^*$ up to this equivalent class, we search for the one 
that sum to zero, i.e. $\sum_{j_1\in[d_1],j_2\in[d_2]} \Theta^*_{j_1,j_2}=0$. 
Let $\bb=\max_{j_1,j_1' \in [d_1],j_2,j_2'\in[d_2]} |\Theta^*_{j_1,j_2}-\Theta^*_{j_1',j_2'}|$, denote the dynamic range of the underlying $\Theta^*$, such that when $k_1\times k_2$ alternatives are presented, we always have
\begin{eqnarray}
	\frac{1}{k_1k_2}e^{-\bb} \;\leq\; \prob{(u_i,v_i)=(j_1,j_2)} \;\leq\; \frac{1}{k_1k_2}e^{\bb}\;,
\end{eqnarray}
for all $(j_1,j_2)\in S_i\times T_i$ and for all $S_i \subseteq [d_1]$ and $T_i \subseteq [d_2]$ such that $|S_i|=k_1$ and $|T_i|=k_2$. We do not make any assumptions on $\bb$ other than that $\bb=O(1)$ with respect to $d_1$ and $d_2$. 
 Assuming $\Theta^*$ is well approximate by a low-rank matrix, we solve the following convex relaxation, given the observed bundled purchase history $\{(u_i,v_i,S_i,T_i)\}_{i\in[n]}$:  
 \begin{eqnarray}
 	\hTheta &\in& \arg\min_{\Theta\in\Omega'_\bb} \calL(\Theta) + \lambda \nucnorm{\Theta} \;,
 	\label{eq:bundleopt}
 \end{eqnarray}
 where the (negative) log likelihood function according to \eqref{eq:defbundleMNL} is 
\begin{eqnarray}
	\cL(\Theta) &=& -\frac{1}{n} \sum_{i=1}^{n} \left( \llangle \Theta, e_{u_i} e_{v_i }^\top    \rrangle - \log \left( \sum_{j_1 \in S_i,j_2\in T_i } \exp\left( \llangle \Theta, e_{j_1} e_{j_2}^\top \rrangle \right) \right) \right),\text{ and }\\
	\label{eq:defbundleL}
	\Omega_{\bb}' &\equiv & \Big\{ A \in \reals^{d_1 \times d_2}   \,\big|\, \lnorm{A}{\infty} \leq \bb \text{, and }\sum_{j_1\in[d_1],j_2\in[d_2]} A_{j_1,j_2}=0 \Big\} \;. 
	\label{eq:defbundleomega}
\end{eqnarray}

Compared to collaborative ranking, 
$(a)$ rows and columns of $\Theta^*$ correspond to an alternative from the first and second category, respectively; 
$(b)$ each sample corresponds to the purchase choice of a user which follow the MNL model with $\Theta^*$; 
$(c)$ each person is presented subsets $S_i$ and $T_i$ of items from each category; 
$(d)$ each sampled data represents the most preferred bundled pair of alternatives.

\subsection{Performance guarantee}
We provide an upper bound on the error achieved by our convex relaxation, 
when the {\em multi-set} of alternatives $S_i$ from the first category  and $T_i$ from the second category 
are drawn uniformly at random with replacement from $[d_1]$ and $[d_2]$ respectively. Precisely, for given $k_1$ and $k_2$, we let $S_i = \{ j_{1,1}^{(i)},\ldots,j_{1,k_1}^{(i)} \}$ 
and $T_i=\{j_{2,1}^{(i)},\ldots, j_{2,k_2}^{(i)}\}$, where 
$j^{(i)}_{1,\ell}$'s and $j^{(i)}_{2,\ell}$'s are independently drawn uniformly at random over 
the $d_1$ and $d_2$ alternatives, respectively. Similar to the previous section, this 
sampling with replacement is necessary for the analysis. 
Define  
\begin{eqnarray}
	\lambda_1=   \sqrt{\frac{e^{2\bb} \max\{d_1,d_2\} \log d}{n\,d_1\,d_2}} \;. 
	\label{eq:bundle_deflambda}
\end{eqnarray} 
\begin{thm}
\label{thm:bundle_ub}
Under the described sampling model, assume $16e^{2\bb} \min\{d_1,d_2\} \log d  \, \leq n \leq \min\{ d^5 ,k_1k_2 \max\{d_1^2,d_2^2\}   \}\log d $, and
$\lambda \in [8 \lambda_1, c_1 \lambda_1]$
with  any constant $c_1=O(1)$ larger than $\max\{8,128/\sqrt{\min\{k_1,k_2\}}\}$. Then, solving the optimization \eqref{eq:bundleopt} achieves 
\begin{eqnarray}
		\frac{1}{d_1d_2} \fnorm{\hTheta-\Theta^\ast}^2  \; \leq \; 
		48 \sqrt{2}  \, e^{2\bb}  c_1 \lambda_1 \sqrt{r} \,  \fnorm{ \hTheta-\Theta^\ast } + 48 e^{2\bb}   c_1 \lambda_1 \sum_{j=r+1}^{\min\{d_1,d_2\}} \sigma_j(\Theta^*) \;, 
	\label{eq:bundle_ub}
\end{eqnarray}
for any $r\in\{1,\ldots, \min\{d_1,d_2\}\}$ 
with probability at least $1-2d^{-3}$ where $d=(d_1+d_2)/2$.
\end{thm}
A proof  is provided in Appendix \ref{sec:bundle_ub_proof}. 
Optimizing over $r$ gives the following corollaries. 
\begin{coro}[{\bf Exact low-rank matrices}]
	Suppose $\Theta^*$ has rank at most $r$. 
	Under the hypotheses of Theorem \ref{thm:bundle_ub}, 
	solving the optimization \eqref{eq:bundleopt} with the choice of the regularization parameter 
	$\lambda\in[8\lambda_1 , c_1\lambda_1 ]$ achieves 	with probability at least $1-2d^{-3}$, 
	\begin{eqnarray}
	\frac{1}{\sqrt{d_1d_2}} \fnorm{\widehat{\Theta}-\Theta^\ast} \,\leq\, 
	48 \sqrt{2}   e^{3 \bb}  c_1 \sqrt{\frac{r(d_1 + d_2) \log d }{n }} \;.  
	\label{eq:bundle_lowrank}
	\end{eqnarray}
	\label{cor:bundle_lowrank}
\end{coro}
This corollary shows that the number of samples $n$ needs to scale as $O(r(d_1+d_2)\log d)$ in order to achieve an arbitrarily small error. This is only a logarithmic factor larger than the degrees of freedom. We provide a fundamental  lower bound on the error, that matches the upper bound up to a logarithmic factor. 
For approximately low-rank matrices in an $\ell_1$-ball 
as defined in \eqref{eq:defBq}, we show 
an upper bound on the error, whose error exponent reduces from one to $(2-q)/2$.

\begin{coro}[{\bf Approximately low-rank matrices}]
	Suppose $\Theta^* \in \bB_q(\rho_q)$ for some $q\in(0,1]$ and $\rho_q>0$. 
	Under the hypotheses of Theorem \ref{thm:bundle_ub}, 
	solving the optimization \eqref{eq:bundleopt} with the choice of the regularization parameter 
	$\lambda \in [8\lambda_1 , c_1\lambda_1]$ achieves 
		with probability at least $1-2d^{-3}$, 
	\begin{eqnarray}
	\frac{1}{\sqrt{d_1d_2}} \fnorm{\widehat{\Theta}-\Theta^\ast} \,\leq\, 
	 \frac{2\sqrt{\rho_q}}{\sqrt{d_1d_2}}  \left(  48 \sqrt{2}  c_1 e^{3 \bb} \,\sqrt{\frac{ d_1d_2(d_1+d_2)\log d }{n }} \right)^{\frac{2-q}{2}}\;. 
	\label{eq:bundle_appxlowrank}
	\end{eqnarray}
	\label{cor:bundle_appxlowrank}
\end{coro}
Since the proof is almost identical to the proof of Corollary \ref{cor:kwise_appxlowrank} in Appendix 
\ref{sec:kwise_cor_proof}, we omit it.


\begin{thm}
	\label{thm:bundle_lb}
Suppose $\Theta^*$ has rank $r$. Under the described sampling model, 
there is a universal constant $c>0$ such that 
\begin{eqnarray}
	\inf_{\hTheta} \sup_{\Theta^*\in\Omega_\bb} \E\Big[ \frac{1}{\sqrt{d_1d_2}}\fnorm{\hTheta-\Theta^*} \Big] &\geq& c\,\min\left\{ \sqrt{\frac{ e^{-5\bb}\,r\,(d_1+d_2)}{n}}  \,,\,  \frac{\bb (d_1+d_2)}{\sqrt{d_1d_2 \log d}} \right\} \;,
\end{eqnarray}
where the infimum is taken over all measurable functions over the observed purchase history 
$\{(u_i,v_i,S_i,T_i)\}_{i\in[n]}$. 
\end{thm}
A proof is provided in Appendix \ref{sec:bundle_lb_proof}.
The first term is the dominant term, and  
when the sample size is comparable to the latent dimension of the problem, 
this theorem establishes that Theorem \ref{thm:bundle_ub} is minimax optimal up to a logarithmic factor.

\section{Discussion}
\label{sec:discussion}

We list remaining challenges for future research. 
$(a)$ Nuclear norm minimization, while polynomial-time, is still slow. 
We want first-order methods that are efficient with provable guarantees. The main challenge is providing a good initialization 
to start such non-convex approaches. $(b)$ For simpler models, such as the PL model, more general sampling over a graph has been studied. We want analytical results for more general sampling. 






{
\small
\bibliographystyle{unsrt}
\bibliography{ranking}
}

\newpage 
\appendix
\section*{Appendix}
\section{Proof of Theorem \ref{thm:kwise_ub}} 
\label{sec:kwise_ub_proof}

We first introduce some additional notations used in the proof. Recall that $\cL(\Theta)$
is the log likelihood function. Let $\nabla \cL(\Theta) \in \reals^{d_1 \times d_2}$
denote its gradient such that $ \nabla_{ij} \cL(\Theta) = \frac{\partial \cL(\Theta)}{\partial \Theta_{ij} }$.
Let $\nabla^2\cL(\Theta) \in \reals^{d_1d_2 \times d_1 d_2}$ denote its Hessian matrix such that
$ \nabla^2_{ij, i'j'} \cL(\Theta) = \frac{\partial^2 \cL(\Theta)}{\partial \Theta_{ij} \partial \Theta_{i'j'} }$.
By the definition of $\cL(\Theta)$ in \eqref{eq:defkwiseL}, we have
\begin{eqnarray}
	\nabla \cL(\Theta^\ast) &=& -\frac{1}{k\,d_1} \sum_{i=1}^{d_1}  \sum_{\ell=1}^{k} e_{i} (e_{v_{i,\ell}}  - p_{i,\ell } )^\top\;,
\end{eqnarray}
where $p_{i,\ell}$ denotes the conditional choice probability at $\ell$-th position. 
Precisely, $p_{i,\ell} = \sum_{j \in S_{i,\ell} }  p_{j|(i,\ell)} e_j$ where 
$p_{j|(i,\ell)}$ is the probability that item $j$ is chosen at $\ell$-th position from the top by the user $i$ 
conditioned on the top $\ell-1$ choices such that   
$p_{j|(i,\ell)} \equiv \prob{v_{i,\ell}  = j | v_{i,1},\ldots, v_{i,\ell-1},S_{i} } = 
e^{\Theta^*_{ij}}/(\sum_{j'\in S_{i,\ell}} e^{\Theta_{ij'}})$ and 
$S_{i,\ell} \equiv S_i\setminus\{v_{i,1},\ldots,v_{i,\ell-1}\}$, 
where $S_i$ is the set of alternatives presented to the $i$-th user and 
 $v_{i,\ell}$ is the item ranked at the $\ell$-th position by the user $i$.
Notice that for $i \neq i'$, $ \frac{\partial^2 \cL(\Theta)}{\partial \Theta_{ij} \partial \Theta_{i'j'} } =0$
and the Hessian is 
\begin{eqnarray}
\frac{\partial^2 \cL(\Theta)}{\partial \Theta_{ij} \partial \Theta_{ij'} }  &= &
	   \frac{1}{k\,d_1} \sum_{\ell=1}^{k} \ind\big(j \in S_{i, \ell} \big) \frac{\partial p_{j|(i,\ell)} }{\partial \Theta_{ij'}} \nonumber\\
&=&  \frac{1}{k\,d_1} \sum_{\ell=1}^{k} \ind\big(j, j' \in S_{i, \ell} \big) \left( p_{j|(i,\ell)} \ind(j=j')  - p_{j|(i,\ell)} p_{j'|(i,\ell)}  \right).
\end{eqnarray}
This Hessian matrix is a block-diagonal matrix $\nabla^2\cL(\Theta)= \diag(H^{(1)}(\Theta), \ldots, H^{(d_1)}(\Theta) )$ with
\begin{align}
H^{(i)}(\Theta) = \frac{1}{k\,d_1} \sum_{\ell=1}^k
 \big(\diag(p_{i,\ell} ) - p_{i,\ell }  p^\top_{i, \ell} \big) \; . \label{eq:DefHessian}
\end{align}

Let $\Delta=\Theta^*-\widehat{\Theta}$ where $\widehat{\Theta}$ is the optimal solution of the convex program in \eqref{eq:kwiseopt}.
We first introduce three key technical lemmas. 
The first lemma follows from Lemma 1 of \cite{NW11},  and
shows that $\Delta$ is approximately low-rank. 

\begin{lemma}
\label{lmm:kwise_deltabound}
If $\lambda\geq 2 \lnorm{\nabla\calL (\Theta^*)}{2}$, then we have
\begin{eqnarray}
	\nucnorm{\Delta} &\le& 4\sqrt{2 r} \fnorm{\Delta} + 4 \sum^{\min\{d_1,d_2\}}_{j = \rho+1}\sigma_j(\Theta^*) \;, \label{eq:kwise_deltabound}
\end{eqnarray}
for all $\rho \in[\min\{d_1,d_2\}]$.
\end{lemma}

The following lemma provides a bound on the gradient
using the concentration of measure for sum of independent random matrices \cite{Jo11}. 
\begin{lemma}\label{lmm:kwise_gradient2}
	For any positive constant $c \ge 1$ and $\log d \ge 4(1+c)/9$, with probability at least $1-2 d^{-c}$,
	\begin{eqnarray}
		\|\nabla \calL(\Theta^\ast)\|_2 &\leq&
		\sqrt{\frac{ 4(1+c)  \, \log d} { k\,d_1^2  }} \max \left\{ \sqrt{d_1/d_2}   \,,\,   e^{2\bb} \sqrt{4(1+c) \log d} \right\}
		\;.
	\end{eqnarray}
\end{lemma}
Since we are typically interested in the regime where 
the number of samples is much smaller than the dimension $d_1\times d_2$ of the problem, 
the Hessian is typically not positive definite. However, when we restrict our attention to the vectorized $\Delta$ with 
relatively small nuclear norm, then we can prove restricted strong convexity, which gives the following bound. 
\begin{lemma}[{\bf Restricted Strong Convexity for collaborative ranking}]
\label{lmm:kwise_hessian2}
Fix any $\Theta \in \Omega _\bb$ and assume $ 24 \,\leq k \leq \min\{ d_1^2, (d_1^2+d_2^2)/(2d_1)\} \log d $.
	Under the random sampling model of the alternatives $\{j_{i\ell}\}_{i\in[d_1],\ell\in[k]}$ and
	the random outcome of the comparisons described in section \ref{sec:intro},
	with probability larger than $1-2d^{-2^{18}}$,
	\begin{eqnarray}
		{\rm Vec}(\Delta)^\top \,\nabla^2\calL(\Theta)\, {\rm Vec}(\Delta)  &\geq& \frac{e^{-4\bb}}{24\,d_1d_2} \fnorm{\Delta}^2\;,
		\label{eq:kwise_hessian2}
	\end{eqnarray}
	for all $\Delta$ in $\calA$ where
	\begin{eqnarray}
		\calA = \Big\{ \Delta \in \reals^{d_1\times d_2} \,\big|\, \lnorm{\Delta}{\infty} \leq 2\bb\,, \, \sum_{j\in[d_2]}\Delta_{ij} = 0\text{ for all }i\in[d_1] \text{ and } \fnorm{\Delta}^2 \geq \mu \nucnorm{\Delta}
		\Big\} \;. \label{eq:kwise_defA}
	\end{eqnarray}
	with 
	\begin{eqnarray}
		\mu &\equiv& 2^{10}\,e^{2\bb}\, \bb\, d_2 \sqrt{\frac{d_1 \, \log d}{k\,\min\{d_1,d_2\} }} \;.
	\label{eq:defmu}
	\end{eqnarray}
\end{lemma}

Building on these lemmas, the proof of Theorem \ref{thm:kwise_ub} is divided into the following two cases.
In both cases, we will show that 
\begin{align}
	\fnorm{\Delta}^2  \;\leq\;  72 \, e^{4\bb}  c_0 \lambda_0  \,d_1d_2\,  \nucnorm{\Delta} \;, 
	\label{eq:errorfrobeniusbound}
\end{align}
 with high probability. Applying Lemma \ref{lmm:kwise_deltabound} 
proves the desired theorem. We are left to show Eq. \eqref{eq:errorfrobeniusbound} holds.

\bigskip
\noindent{\bf Case 1: Suppose $\fnorm{\Delta}^2 \geq  \mu \,\nucnorm{\Delta} $.}
With $\Delta=\Theta^* - \hTheta$, the Taylor expansion yields
\begin{align}
\calL(\widehat{\Theta})=\calL(\Theta^\ast) -  {\llangle \nabla \calL(\Theta^\ast), \Delta \rrangle} + \frac{1}{2} {\rm Vec}(\Delta) \nabla^2\calL(\Theta) {\rm Vec}^\top (\Delta),
\label{eq:LikelihoodTaylor}
\end{align}
where $\Theta=a \widehat{\Theta} + (1-a) \Theta^\ast$ for some $a \in [0,1]$.
It follows from Lemma \ref{lmm:kwise_hessian2} that with probability at least $1-2d^{-2^{18}}$,
\begin{align*}
	 \calL(\widehat{\Theta}) -\calL(\Theta^\ast) & \;\ge\;  -\llangle \nabla \calL(\Theta^\ast), \Delta \rrangle
	+ \frac{ e^{-4\bb}}{48\,d_1\,d_2} \fnorm{\Delta}^2\\
	&\; \ge\; - \lnorm{\nabla\calL(\Theta^\ast)}{2} \nucnorm{\Delta}+ \frac{  e^{-4\bb}}{48\,d_1\,d_2} \fnorm{\Delta}^2\;.
\end{align*}
From the definition of $\widehat{\Theta}$ as an optimal solution of the minimization, we have
\begin{align*}
\calL(\widehat{\Theta})-  \calL(\Theta^\ast)  \;\le \; \lambda \left(  \nucnorm{\Theta^\ast} - \nucnorm{\widehat{\Theta}} \right) \;\le\; \lambda \nucnorm{\Delta}\;.
\end{align*}
By the assumption, we choose $\lambda\geq 32\lambda_0$. 
 In view of Lemma \ref{lmm:kwise_gradient2}, this implies that 
  $\lambda \geq 2\lnorm{\nabla\calL (\Theta^*)}{2} $ 
  with probability at least $1-2d^{-3}$. 
   It follows that with probability at least $1-2d^{-3}- 2d^{-2^{18}}$,
\begin{align*}
\frac{ e^{-4\bb}}{48d_1 d_2 } \fnorm{\Delta}^2 \;  \leq \;  \big(\lambda + \lnorm{\nabla\calL(\Theta^*)}{2}\big)\,  \nucnorm{\Delta} \;\leq\; \frac{3 \lambda}{2}  \nucnorm{\Delta} \;.
\end{align*}
By our assumption on $\lambda \leq c_0\lambda_0$, this proves the desired bound in Eq. \eqref{eq:errorfrobeniusbound}

\noindent{\bf Case 2:  Suppose $\fnorm{\Delta}^2 \leq  \mu \,\nucnorm{\Delta} $.}  
By the definition of $\mu$ and the fact that $c_0 \ge 32$, it follows that 
$\mu \le 72 \, e^{4\bb}  c_0 \lambda_0  \,d_1d_2$, and we get the same bound as in Eq. \eqref{eq:errorfrobeniusbound}.

\subsection{Proof of Lemma \ref{lmm:kwise_deltabound}}
Denote the singular value decomposition of $\Theta^\ast$ by $\Theta^\ast = U \Sigma V^\top$, where $U\in \reals^{d_1 \times d_1}$ and $V \in \reals^{d_2 \times d_2}$
 are orthogonal matrices.
 For a given $r\in[\min\{d_1,d_2\}]$, 
 Let $U_r=[u_1, \ldots, u_r]$  and $V_r=[v_1, \ldots, v_r]$, where $u_i \in \reals^{d_1 \times 1}$ and $v_i \in \reals^{d_2 \times 1}$ are the left and right singular vectors corresponding to the $i$-th largest singular value, respectively.
Define $T$ to be the subspace spanned by all matrices in $\reals^{d_1 \times d_2}$ of the form $U_rA^\top$ or $BV_r^\top$ for any $A\in \mathbb{R}^{d_2\times r}$ or
$B \in \reals^{d_1 \times r}$, respectively.
The orthogonal projection of any matrix $M\in \mathbb{R}^{d_1 \times d_2}$ onto the space $T$ is given by $\mathcal{P}_T(M) = U_rU_r^\top M+MV_rV_r^\top-U_rU_r^\top MV_rV_r^\top $. The projection of $M$ onto the complement space $T^\perp$ is $\mathcal{P}_{T^\perp}(M) = (I-U_rU_r^\top) M (I-V_r V_r^\top )$.
The subspace $T$ and the respective projections onto $T$ and $T^\perp$ play crucial a role in the analysis of nuclear norm minimization, since they define the sub-gradient of the nuclear norm at $\Theta^*$. We refer to \cite{CR09} for more detailed treatment of this topic. 

Let $\Delta'=\calP_{T}(\Delta)$ and $\Delta'' = \calP_{T^\perp} (\Delta)$. Notice that $\calP_{T} (\Theta^\ast) = U_r \Sigma_r V_r^\top$, where
$\Sigma_r \in \reals^{r \times r}$ is the diagonal matrix formed by the top $r$ singular values. Since $\calP_{T} (\Theta^\ast)$ and $\Delta''$
have row and column spaces that are orthogonal, it follows from Lemma 2.3 in \cite{RFP10} that
\begin{align*}
\nucnorm{ \calP_{T} (\Theta^\ast) - \Delta''} =\nucnorm{\calP_{T} (\Theta^\ast)} + \nucnorm{\Delta''}\;.
\end{align*}
Hence, in view of the triangle inequality,
\begin{align}
\nucnorm{\hTheta}&= \nucnorm{ \calP_{T} (\Theta^\ast) + \calP_{T^\perp} (\Theta^\ast) - \Delta' - \Delta''} \nonumber \\
& \ge \nucnorm{ \calP_{T} (\Theta^\ast)- \Delta'' }  - \nucnorm{\calP_{T^\perp} (\Theta^\ast) - \Delta' }\nonumber \\
& =  \nucnorm{ \calP_{T} (\Theta^\ast)} + \nucnorm{\Delta''} -   \nucnorm{\calP_{T^\perp} (\Theta^\ast) - \Delta' } \nonumber \\
& \ge \nucnorm{ \calP_{T} (\Theta^\ast)} + \nucnorm{\Delta''}  - \nucnorm{\calP_{T^\perp} (\Theta^\ast) } - \nucnorm{\Delta'} \nonumber \\
& =  \nucnorm{\Theta^\ast} + \nucnorm{\Delta''}  -2 \nucnorm{\calP_{T^\perp} (\Theta^\ast) }- \nucnorm{\Delta'}. \label{eq:nucleardecomp}
\end{align}
Because $\widehat{\Theta}$ is an optimal solution, we have
\begin{align}
	\lambda \left( \nucnorm{\widehat{\Theta}}- \nucnorm{\Theta^\ast} \right)  \le 
	 \calL(\Theta^\ast) - \calL(\widehat{\Theta})  \overset{(a)}{\le}  
	\Iprod{\Delta}{\nabla \calL(\Theta^\ast) } \overset{(b)}{\le} 
	\nucnorm{\Delta}  \lnorm{\nabla \calL(\Theta^\ast)}{2}  \le \frac{\lambda}{2} \nucnorm{\Delta},   \label{eq:optimalitycondition}
\end{align}
where $(a)$ holds due to the concavity of $\calL$; $(b)$ follows from the Cauchy-Schwarz inequality; the last inequality holds due to
the assumption that $\lambda \geq 2 \lnorm{\nabla\calL (\Theta^*)}{2}$. Combining \eqref{eq:nucleardecomp} and \eqref{eq:optimalitycondition} yields
\begin{align*}
2 \left(  \nucnorm{\Delta''}  -2 \nucnorm{\calP_{T^\perp} (\Theta^\ast) }- \nucnorm{\Delta'} \right) \le \nucnorm{\Delta} \le  \nucnorm{\Delta'} +  \nucnorm{\Delta''}.
\end{align*}
Thus $  \nucnorm{\Delta'' } \le 3  \nucnorm{\Delta' } + 4  \nucnorm{ \calP_{T^\perp} (\Theta^\ast) }$. By triangle inequality, 
\begin{eqnarray*}
	\nucnorm{\Delta } \le 4\nucnorm{\Delta' } + 4  \nucnorm{ \calP_{T^\perp} (\Theta^\ast) }\;.
\end{eqnarray*} 
 Notice that $\Delta'=U_r U_r^\top \Delta+ ( I-U_rU_r^\top)  \Delta V_rV_r^\top$. Both $U_r U_r^\top \Delta$ and  $ (I-U_rU_r^\top)  \Delta V_rV_r^\top$ have rank at most $r$.  Thus  $\Delta'$ has rank at most $2r$. Hence, $\nucnorm{\Delta' }
\le \sqrt{2r} \fnorm{\Delta'} \le \sqrt{2r} \fnorm{\Delta}$. Then the theorem follows because $\nucnorm{ \calP_{T^\perp} (\Theta^\ast) } = \sum^{\min\{d_1,d_2\}}_{j=r+1}\sigma_j(\Theta^*).$

\subsection{Proof of Lemma \ref{lmm:kwise_gradient2}}

Define $X_i=  -e_{i} \sum_{\ell=1}^{k} (e_{v_{i,\ell} } - p_{i,\ell } )^\top $ such that $\nabla \calL(\Theta^\ast)= \frac{1}{k\, d_1} \sum_{i=1}^{d_1} X_i$, which is a sum of $d_1$ independent random matrices. Note that
since $ p_{i,\ell }$ has $(k+1-\ell)$ non-zero entries, each bounded in absolute value by $e^{2\bb}/(k+1-\ell)$,
we have the following bound deterministically:
\begin{eqnarray*}
	\lnorm{X_i}{2}  &=& \Big\| \sum_{\ell=1}^{k} \big(e_{v_{i,\ell} } - p_{i,\ell } \big) \Big\|\\
	&\le&\sqrt{k} +   \sum_{\ell=1}^{k}  \Big\| p_{i,\ell }  \Big\| \\
	&\leq& \sqrt{k} + e^{2\bb} \sum_{\ell=1}^k\frac{1}{\sqrt{k+1-\ell}}\\
	&\leq& \sqrt{k} + e^{2\bb} 2 (\sqrt{k+1} - 1)\\
	&\leq& 3 e^{2\bb} \sqrt{k}\;,
\end{eqnarray*}
and
\begin{align*}
\lnorm{\sum_i \expect{X_i X_i^\top}}{2} & \leq 9 e^{4\bb} k \lnorm{\sum_{i=1}^{d_1} \expect{e_{i} e_{i}^\top}}{2} =\,9 e^{4\bb}k \lnorm{{\mathbb I}_{d_1\times d_1}}{2} =\, 9 e^{4\bb} k,
\end{align*}
and
\begin{align*}
\sum_{i=1}^{d_1} \expect{X_i^\top X_i} & = \sum_{i=1}^{d_1} \sum_{\ell,\ell'=1}^{k} \expect{(e_{i,\ell} - p_{i,\ell }) (e_{v_{i,\ell'} } - p_{i, \ell' })^\top  } \\
  & = \sum_{i=1}^{d_1} \sum_{\ell=1}^{k} \expect{(e_{i,\ell} - p_{i,\ell }) (e_{v_{i,\ell} } - p_{i, \ell })^\top  } \\
& =  \sum_{i=1}^{d_1} \sum_{\ell=1}^{k} \left( \expect{ e_{v_{i,\ell} } e_{v_{i,\ell} }^\top } - \expect{p_{i,\ell } p^\top_{i,\ell} } \right)  \\
& \preceq \sum_{i=1}^{d_1} \sum_{\ell=1}^{k} \expect{ e_{v_{i,\ell} } e_{v_{i,\ell} }^\top } \\
& = \sum_{i=1}^{d_1}   \frac{k}{d_2} {\mathbb I}_{d_2 \times d_2} \; .
\end{align*}
Therefore,
\begin{align*}
	\lnorm{\sum_{i=1}^{d_1} \expect{X_i^\top X_i} }{2} & \leq \frac{k d_1}{d_2} \;.
\end{align*}
By matrix Bernstein inequality \cite{Jo11},
\begin{align*}
	{\mathbb P} \Big( \lnorm{\nabla \calL(\Theta^\ast)}{2} > t \Big) \leq (d_1+d_2)
	\exp \Big( \frac{- k^2\,d_1^2 \, t^2/2}{ (d_1k/\min\{d_2,d_1/(9e^{4\bb})\} )+ (3e^{2\bb} k^{3/2} d_1 t/3)} \Big)\;,
\end{align*}
which gives the desired tail probability of $2d^{-c}$ for the choice of
\begin{eqnarray*}
	t&=& \max \left\{  \sqrt{\frac{4(1+c) \, \log d}{ k\,d_1\, \min\{d_2,d_1/(9e^{4\bb})\} }} \,,\,  \frac{4(1+c)e^{2\bb} \log d}{ k^{1/2} \,d_1}\right\} \\
& =& \frac{  \sqrt{ 4(1+c)  \, \log d}  } { k^{1/2}\, d_1  } \max \left\{ \sqrt{d_1/d_2}   \,,\,  e^{2\bb} \sqrt{4(1+c) \log d} \right\},
\end{eqnarray*}
where the last equality holds due to the assumption that $ \log d \ge 4(1+c)/ 9$.

\subsection{Proof of Lemma \ref{lmm:kwise_hessian2}}

Recall that the Hessian matrix is a block-diagonal matrix with the $i$-th block
$H^{(i)}(\Theta)$ given by \eqref{eq:DefHessian}. 
We use the following remark from \cite{HOX14} to bound the Hessian. 
\begin{rem}   
	 \cite[Claim 1]{HOX14}
	Given $\theta \in   \reals^r,$  let 
	$p$ be the column probability vector with $p_i=e^{\theta_i}/(e^{\theta_1}+\cdots + e^{\theta_\rho})$ for each $i\in[\rho]$  
	and for any positive integer $\rho$. 
	If  $| \theta_i | \leq \bb,$ for all $i \in [\rho]$,  then 
	$$e^{2\bb} \Big(\diag{(p)} - pp^T\Big) \;\succeq\;  \frac{1}{\rho} \diag(\mathbf{1}) - \frac{1}{\rho^2}\mathbf{1}\mathbf{1}^\top  \;.$$
	\label{rem:hess}
\end{rem}
By letting $\mathbf{1}_{S_{i,\ell}} = \sum_{j \in S_{i,\ell} } e_j$ and applying the above claim, we have
\begin{align*}
  e^{2\alpha} H^{(i)}(\Theta) & \succeq \frac{1}{k\, d_1} \sum_{\ell=1}^k  \left(  \frac{1}{k-\ell+1 }\diag(\mathbf{1}_{S_{i,\ell}}) - \frac{1}{(k-\ell+1)^2 }  \mathbf{1}_{S_{i,\ell}}\mathbf{1}^\top_{S_{i,\ell}} \right) \\
 &= \frac{1}{2\,k\,d_1}  \sum_{\ell=1}^k \frac{1}{(k-\ell+1)^2}\sum_{j, j' \in S_{i,\ell}} (e_j-e_{j'})(e_j-e_{j'})^\top \\
 & \succeq \frac{1}{2\,k^3\, d_1} \sum_{\ell=1}^k \sum_{j, j' \in S_{i,\ell}} (e_j-e_{j'})(e_j-e_{j'})^\top.
\end{align*}
Hence,
\begin{align*}
 {\rm Vec}(\Delta) \nabla^2\calL(\Theta) {\rm Vec}^\top (\Delta)  &=  \sum_{i=1}^{d_1} (\Delta^\top e_i)^\top H^{(i)} (\Theta) (\Delta^\top e_i ) \\
 & \ge   \frac{e^{-2\alpha}}{2\,k^3\,d_1 } \sum_{i=1}^{d_1} \sum_{\ell=1}^k \sum_{j,j' \in S_{i,\ell} } \lnorm{e_i^\top \Delta (e_j - e_{j'} )}{2}^2.
\end{align*}
By changing the order of the summation, we get that
\begin{align*}
\sum_{\ell=1}^k \sum_{j,j' \in S_{i,\ell} } \lnorm{e_i^\top \Delta (e_j - e_{j'} )}{2}^2 = \sum_{ \ell, \ell'=1}^{k} \Iprod{\Delta}{e_{i, j_{i,\ell}} -e_{i, j_{i,\ell'} } }^2 \sum_{\ell''=1}^k \ind\big(\,\sigma_i(j_{i,\ell^{''}}) \le \min \{\sigma_i(j_{i,\ell}), \sigma_i(j_{i,\ell'} )   \}\,\big).
\end{align*}
Define 
\begin{eqnarray}
	\label{eq:defchi}
	\chi_{i,\ell,\ell',\ell^{''}} &\equiv& \ind\big(\,\sigma_i(j_{i,\ell^{''}}) \le \min \{\sigma_i(j_{i,\ell}), \sigma_i(j_{i,\ell'} )   \}\,\big)\;,
\end{eqnarray}
 and let 
\begin{align*}
	H(\Delta ) \; \equiv \;  \frac{e^{-2\alpha}}{2\,k^3\,d_1 } \sum_{i=1}^{d_1} \sum_{ \ell, \ell'=1}^{k} \Iprod{\Delta}{e_{i, j_{i,\ell}} -e_{i, j_{i,\ell'} } }^2 \sum_{\ell''=1}^k \chi_{i,\ell,\ell',\ell^{''}}  .
\end{align*}
Then we have ${\rm Vec}^\top(\Delta) \nabla^2\calL(\Theta) {\rm Vec} (\Delta) \ge H(\Delta)$. To prove the theorem, it suffices to bound $H(\Delta)$ from the below. 
First, we prove a lower bound on  the expectation $\E[H(\Delta)]$.
Notice that for $\ell\neq \ell'$, the conditional expectation of $\chi_{i,\ell,\ell',\ell''}$'s, given the set of alternatives presented to user $i$ is 
\begin{align*}
  {\mathbb E} \Big[ \sum_{\ell''=1}^k \chi_{i,\ell,\ell',\ell^{''}} \,\big| \, j_{i,1}, \ldots, j_{i,k}\Big]  &= 1+ \sum_{\ell'' \neq \ell, \ell'} \frac{\exp( \theta_{i,j_{i,\ell''}} ) } { \exp( \theta_{i,j_{i,\ell''}} ) +\exp( \theta_{i,j_{i,\ell'}} ) +\exp( \theta_{i,j_{i,\ell}} )}  \\
&\ge 1+ \frac{k-2}{1+2e^{2\bb}} \ge \frac{k}{3 e^{2\bb}}.
\end{align*}
Then,
\begin{eqnarray}
	\E[H(\Delta)] &=&
	\frac{e^{-2\bb}}{2\,k^3\,d_1}  \sum_{i,\ell,\ell'}
{\mathbb E}\Big[\llangle \Delta, e_{i,j_{i,\ell}}-e_{i,j_{i,\ell'}}\rrangle^2  {\mathbb E} \big[ \sum_{\ell''=1}^{k} \chi_{i,\ell,\ell',\ell^{''}} \,\big| \, j_{i,1}, \ldots, j_{i,k} \big] \Big]\nonumber\\
&\geq& \frac{e^{-4\bb}}{6\,k^2\,d_1 }  \sum_{i=1}^{d_1} \sum_{\ell,\ell' \in [k]} \expect{
\llangle \Delta, e_{i,j_{i,\ell}}-e_{i,j_{i,\ell'}}\rrangle^2}   \nonumber\\
&=&  \frac{e^{-4\bb}}{6\,k^2\,d_1 }  \sum_{i=1}^{d_1}
	\sum_{\ell \neq \ell'\in [k]}  \left( \frac{2}{d_2}\sum_{j=1}^{d_2}\Delta_{ij}^2 - \frac{2}{d_2^2}\sum_{j,j'=1}^{d_2} \Delta_{ij}\Delta_{ij'}\right)\nonumber\\
&=& \frac{e^{-4\bb}(k-1)}{3\,k\,d_1\,d_2} \fnorm{\Delta}^2\;,
\label{eq:kwise_hessexp}
\end{eqnarray}
where the last equality holds because  $\sum_{j\in[d_2]}\Delta_{ij}=0$
for $\Delta \in \Omega_{2\bb}$ and for all $i\in[d_1]$.

We are left to prove that $H(\Delta)$ cannot deviate from its mean too much.  
Suppose there exists a
$\Delta \in \calA$ such that Eq. \eqref{eq:kwise_hessian2} is violated, i.e.
$H(\Delta) < (e^{-4 \bb}/(24 \,d_1d_2)) \fnorm{\Delta}^2$. 
We will show this happens with a small probability. 
From Eq. \eqref{eq:kwise_hessexp}, we  get that  for $k\geq 24$, 
\begin{eqnarray}
	\E[H(\Delta)] - H(\Delta) &\geq& \frac{(7k-8)}{24k} \frac{ e^{-4\bb} }{d_1\, d_2} \fnorm{\Delta}^2 \nonumber\\
		&\geq&\frac{ (20/3)\, e^{-4\bb}  }{24 \,d_1 d_2} \fnorm{\Delta}^2 \;. \label{eq:kwise_peeling1}
\end{eqnarray}
We use a peeling argument as in \cite[Lemma 3]{NW11}, \cite{Van00}  
to upper bound the probability that Eq. \eqref{eq:kwise_peeling1} is true. 
We first construct the following family of subsets to cover $\calA$ such that 
$\calA \subseteq \bigcup_{\ell=1}^\infty \calS_\ell$. 
Recall  $\mu=2^{10}e^{2\bb} \bb d_2 \sqrt{(d_1\log d)/(k\min\{d_1,d_2\})}$, define in \eqref{eq:defmu}. 
Notice that since for any $\Delta\in \calA$, $\fnorm{\Delta}^2 \ge \mu  \nucnorm{\Delta} \ge \mu \fnorm{\Delta}$, it follows that
$\fnorm{\Delta} \ge \mu$.
Then, we can cover $\calA$ with the family of sets 
\begin{eqnarray*}
	\calS_\ell =\Big\{ \Delta\in\reals^{d_1\times d_2} \,\Big|\, \lnorm{\Delta}{\infty} \leq 2\bb \,,\, \beta^{\ell-1}\mu \leq \fnorm{\Delta} \leq \beta^\ell \mu \,,\,  \sum_{j\in[d_2]} \Delta_{ij}=0 \text{ for all }i\in[d_1], \text{ and }  \nucnorm{\Delta} \leq  \beta^{2\ell}\mu 	 \Big\} \;,
\end{eqnarray*}
where $\beta=\sqrt{10/9}$ and for $\ell\in\{1,2,3,\ldots \}$.
This implies that 
when there exists a $\Delta\in\calA$ such that 
\eqref{eq:kwise_peeling1} holds, then there exists an $\ell\in\Z_+$ such that $\Delta\in \calS_\ell$ and 
\begin{eqnarray}
		\E[H(\Delta)] - H(\Delta) &\geq & \frac{ (20/3)\, e^{-4\bb}  }{24\, d_1 d_2} \beta^{2(\ell-1)} \mu^2 \nonumber\\
		&\geq & \frac{e^{-4\bb} }{4 \,d_1 d_2} \beta^{2 \ell} \mu^2 \;.
		\label{eq:kwise_peeling2} 
\end{eqnarray}

Applying the union bound over $\ell\in\Z_+$, we get from \eqref{eq:kwise_peeling1} and \eqref{eq:kwise_peeling2} that 
\begin{align}
	&\prob{\exists \Delta \in \calA \;,\; H(\Delta) < \frac{ e^{-4\bb} } { 24 \,d_1 d_2 } \fnorm{\Delta}^2 }\; \leq\;
		\sum_{\ell=1}^\infty \prob{ \sup_{\Delta\in\calS_\ell} \big(\; \E[ H(\Delta)] - H(\Delta) \;\big) > \frac{e^{-4 \bb} }{ 4 \,d_1 d_2}(\beta^\ell \mu)^2 } \nonumber\\
		&\hspace{3.6cm}\leq\;\; \sum_{\ell=1}^\infty \prob{ \sup_{\Delta\in\calB(\beta^\ell\mu)} \big(\; \E[ H(\Delta)] - H(\Delta) \;\big) > \frac{e^{-4 \bb}  }{ 4 \,d_1 d_2}(\beta^\ell \mu)^2 }
		\;, \label{eq:kwise_peeling3}
\end{align}
where we define a new set $\calB(D)$ such that $\calS_\ell \subseteq \calB(\beta^\ell \mu)$: 
\begin{align}
	\calB(D) = \big\{\, \Delta \in \reals^{d_1\times d_2} \,\big|\, \|\Delta\|_\infty \leq 2\bb, \fnorm{\Delta}\leq D, \sum_{j\in[d_2]} \Delta_{ij}=0 \text{ for all }i\in[d_1],  \mu \nucnorm{\Delta} \leq D^2 \,
		  \Big\} \;. \label{eq:defcB}
\end{align}
The following key lemma provides the upper bound on this probability.  
\begin{lemma}
For $(16\min\{d_1,d_2\} \log d)/(3d_1)\leq k\leq d_1^2 \log d$, 
\begin{eqnarray}
	\prob{\sup_{\Delta\in\calB(D)} \Big(\; \E[ H(\Delta)] - H(\Delta) \;\Big) \geq \frac{e^{-4\bb}}{4d_1d_2} D^2 } &\leq& \exp\Big\{ - \frac{e^{-4\bb} \,k\,D^4}{2^{19} \bb^4 d_1 d_2^2 } \Big\}\;.
	\label{eq:kwise_hessian3}
\end{eqnarray}
\label{lmm:kwise_hessian3}
\end{lemma}

Let $\eta= \exp\left(-\frac{e^{-4\bb}4k(\beta-1.002) \mu^4}{2^{19} \bb^4d_1 d_2^2} \right)$. 
Applying the tail bound to \eqref{eq:kwise_peeling3}, we get 
\begin{eqnarray*}
	\prob{\exists \Delta \in \calA \;,\; H(\Delta) < \frac{ e^{-4\bb} } { 24 \,d_1 d_2 } \fnorm{\Delta}^2 } 
	&\leq & \sum_{\ell=1}^\infty \exp \Big\{-\frac{e^{-4\bb}k(\beta^\ell \mu)^4 }{2^{19} \bb^4 d_1 d_2^2}\Big\} \\
	& \overset{(a)}{\leq} &\sum_{\ell=1}^\infty \exp\Big\{-\frac{e^{-4 \bb}4k\ell (\beta-1.002) \mu^4 }{2^{19} \bb^4d_1 d_2^2}\Big\} \\
	&\leq & \frac{\eta}{1-\eta},
\end{eqnarray*}
where $(a)$ holds because $\beta^{x} \geq x \log\beta \ge x(\beta-1.002)$ for the choice of $\beta=\sqrt{10/9}$.
By the definition of $\mu$,
\begin{align*}
\eta \; = \; \exp\Big\{ - \frac{  2^{23}\,e^{4 \bb} d_2^2 d_1 (\log d)^2 (\beta-1.002) }{k (\min\{d_1,d_2\})^2}  \Big\} \;  \le\;   \exp \{ -\,2^{18}\,\log d\} \;,
\end{align*}
where the last inequality follows from the assumption that 
$k\leq \max\{d_1,d_2^2/d_1\}\log d =  (d_2^2 d_1 \log d) / (\min\{d_1,d_2\})^2 $, 
and $\beta - 1.002\geq 2^{-5}$. 
Since for $d \ge 2$, $\exp\{-2^{18}\log d\} \leq 1/2$ and thus $\eta \le 1/2$, the lemma follows by assembling the last two
displayed inequalities.

\subsection{Proof of Lemma \ref{lmm:kwise_hessian3}}
\label{sec:kwise_hessian3_proof}

Recall that
\begin{align*}
H(\Delta ) = \frac{e^{-2\alpha}}{2\,k^3\,d_1 } \sum_{i=1}^{d_1} \sum_{ \ell, \ell'=1}^{k} \Iprod{\Delta}{e_{i, j_{i,\ell}} -e_{i, j_{i,\ell'} } }^2 \sum_{\ell''=1}^k \chi_{i,\ell,\ell',\ell^{''}}  \;,
\end{align*}
with $\chi_{i,\ell,\ell',\ell^{''}} = \indc{\sigma_i(j_{i,\ell^{''}}) \le \min \{\sigma_i(j_{i,\ell}), \sigma_i(j_{i,\ell'} )   \}}$.
Let $Z = \sup_{\Delta\in\calB(D)}  \E[H(\Delta)] - H(\Delta)$ be the worst-case random deviation of $H(\Delta)$ form its mean.
We prove an upper bound on $Z$ by showing that
$Z - \E[Z] \leq e^{-4 \bb} D^2/(64 d_1d_2) $ with high probability, and $\E[Z] \leq 9 e^{-4 \bb} D^2/(40d_1d_2)$.
This proves the desired claim in Lemma \ref{lmm:kwise_hessian3}.

To prove the concentration of $Z$,
we utilize the random utility model (RUM) theoretic  interpretation of the MNL model.
The random variable $Z$ depends on the random choice of alternatives $\{j_{i,\ell}\}_{i\in[d_1],\ell\in[k]}$
and the random $k$-wise ranking outcomes $\{\sigma_i\}_{i\in[d_1]}$.
The random utility theory, pioneered by  \cite{Thu27,Mar60,Luce59}, 
tells us that the $k$-wise ranking from the MNL model
has the same distribution as first drawing independent
(unobserved) utilities $u_{i, \ell}$'s of the item $j_{i,\ell}$ for user $i$
according to the standard Gumbel Cumulative Distribution Function (CDF) $F(c-\Theta_{i, j_{i,\ell}})$ with $F(c)=e^{-e^{-c} } $,
and then ranking the $k$ items for user $i$ according to their respective utilities.
Given this definition of the MNL model, we have $\chi_{i,\ell,\ell',\ell^{''}} = \indc{u_{i,\ell^{''} }  \ge \max \{ u_{i,\ell } , u_{i,\ell'}    \} }$.
Thus $Z$ is a function of independent choices of the items and their (unobserved) utilities, i.e. $Z=f(\{(j_{i,\ell},u_{i,\ell} ) \}_{i\in[d_1],\ell\in[k]})$.
Let $x_{i,\ell} = (j_{i,\ell}, u_{i,\ell} )$ and write $H(\Delta)$ as $H(\Delta, \{ x_{i,\ell} \}_{i\in[d_1],\ell\in[k]} )$.
This allows us to bound the difference and apply McDiarmid's tail bound. 
Note that for any  $i\in[d_1]$, $\ell\in[k]$,
$x_{1,1} ,\ldots, x_{d_1, k} $, and $x'_{i, \ell} $,
\begin{align*}
& \big|\,f \big( \,x_{1,1}, \ldots,x_{i,\ell},\ldots,x_{d_1, k} \,\big) - f\big(\,x_{1,1} ,\ldots, x'_{i,\ell} ,\ldots, x_{d_1, k} \,\big)\,\big|\\
& =\big| \sup_{\Delta \in \calB(D) } \left( \expect{H(\Delta)} - H(\Delta, x_{1,1}, \ldots,x_{i,\ell},\ldots,x_{d_1, k} ) \right) - \sup_{\Delta \in \calB(D) } \left( \expect{H(\Delta) } - H(\Delta, x_{1,1} ,\ldots, x'_{i,\ell} ,\ldots, x_{d_1, k}) \right) \big|  \\
& \le \sup_{\Delta\in \calB(D)  } \big| H(\Delta, x_{1,1}, \ldots,x_{i,\ell},\ldots,x_{d_1, k} )  -H(\Delta, x_{1,1} ,\ldots, x'_{i,\ell} ,\ldots, x_{d_1, k})   \big| \\
& \overset{(a)}{\le} \frac{e^{-2\bb} }{2\,k^3\, d_1 }  \sup_{\Delta \in \calB(D) } 	\Big\{ 2  \sum_{\ell'\in[k]}
	 \llangle \Delta, e_{i,j_{i,\ell}}-e_{i,j_{i,\ell'}}\rrangle^2  \sum_{\ell''=1}^k \chi_{i,\ell,\ell',\ell^{''}} + \sum_{\ell',\ell''\in[k]}
	\llangle \Delta, e_{i,j_{i,\ell'}}-e_{i,j_{i,\ell''}}\rrangle^2  \chi_{i,\ell',\ell'',\ell } \Big\} \\
& \overset{(b)}{\le} \frac{8 \bb^2 e^{-2\bb} }{  k^3\,d_1 }  	\Big\{ 2  \sum_{\ell'\in[k]\backslash\{\ell\} }
	  \sum_{\ell''=1}^k \chi_{i,\ell,\ell',\ell^{''} } + \sum_{\ell',\ell''\in[k], \ell' \neq \ell'', }
	\chi_{i,\ell',\ell'',\ell } \Big\} \\
& \le \frac{ 16 \bb^2 e^{-2\bb} }{ k\,d_1 } \;,
\end{align*}
where $(a)$ follows because for a fixed  $i$ and $\ell$, the random variable $x_{i,\ell}=(j_{i,\ell},u_{i,\ell})$ can appear 
in three terms, i.e. $\sum_{\ell',\ell''} \llangle \Delta,e_{i,j_{i,\ell}}-e_{i,j_{i,\ell'}}\rrangle^2\chi_{i,\ell,\ell',\ell''}
+ \sum_{\ell',\ell''} \llangle \Delta,e_{i,j_{i,\ell'}}-e_{i,j_{i,\ell}}\rrangle^2\chi_{i,\ell',\ell,\ell''} + \sum_{\ell',\ell''} \llangle \Delta,e_{i,j_{i,\ell'}}-e_{i,j_{i,\ell''}}\rrangle^2\chi_{i,\ell',\ell'',\ell}$, 
and $(b)$ follows because $|\Delta_{ij}|\leq 2\bb$ for all $i$, $j$ since $\Delta\in \calB(D)$. 
The last inequality follows because
in the worst case, 
$\sum_{\ell'\in[k]\backslash\{\ell\}}	  \sum_{\ell''=1}^k \chi_{i,\ell,\ell',\ell^{''} }  \leq k(k-1)/2$ 
and $\sum_{\ell',\ell''\in[k] , \ell' \neq \ell''} \chi_{i,\ell',\ell'',\ell }  \leq  k(k-1)$. 
This holds with equality if $\sigma_i(j_{i,\ell})=k$ and $\sigma_i(j_{i,\ell})=1$, respectively. 
By bounded differences inequality, we have
\begin{align*}
\prob{Z - \expect{Z} \ge t } \,\le\, \exp\left( - \frac{k^2\,d_1^2\, t^2}{ 2^7\, \bb^4e^{-4\bb}  d_1 k}\right),
\end{align*}
It follows that for the choice of $t=e^{-4\bb} D^2/(64 d_1d_2)$,
\begin{align*}
\prob{Z - \expect{Z} \ge \frac{e^{-4\bb} D^2}{ 64 d_1d_2} } \le \exp\Big( - \frac{e^{-4\bb} k D^4 }{2^{19} \bb^4  d_1 d_2^2 } \Big) \;.
\end{align*}

We are left to prove the upper bound on $\E[Z]$ using symmetrization and contraction.
Define random variables 
\begin{eqnarray}
	Y_{i,\ell,\ell',\ell''}(\Delta) &\equiv& (\Delta_{i,j_{i,\ell}} - \Delta_{i,j_{i,\ell'}})^2  \chi_{i,\ell,\ell',\ell''} \;,
	\label{eq:defY}
\end{eqnarray}
 where the randomness is in the choice of
alternatives $j_{i,\ell},j_{i,\ell'},$ and $j_{i,\ell''}$, and the outcome of the comparisons of those three alternatives.

The main challenge in applying the symmetrization to 
$\sum_{\ell,\ell',\ell''\in[k]} Y_{i,\ell,\ell',\ell''}(\Delta) $ is that 
we need to 
partition the summation over the set $[k]\times[k]\times[k]$ into 
subsets of independent random variables, such that we can apply the standard symmetrization argument. 
to this end, we prove in the following lemma a  
a generalization of the well-known problem of 
scheduling a round robin tournament 
to a tournament of matches involving three teams each.   
No teams are present in more than one triple in a single round, and we want to minimize the number of rounds to cover all combination of triples are matched. 
For example, 
when there are $k=6$ teams, there is a simple construction of such a tournament: 
$T_1=\{(1,2,3),(4,5,6)\}$, $T_2=\{1,2,4),(3,5,6)\}$, $T_3=\{(1,2,5),(3,4,6)\}$, 
$T_4=\{(1,2,6),(3,4,5)\}$, $T_5=\{(1,3,4),(2,5,6)\}$, $T_6=\{(1,3,5),(2,4,6)\}$, 
$T_7=\{(1,3,6),(2,4,5)\}$, $T_8=\{(1,4,5),(2,3,6)\}$, $T_9=\{(1,4,6),(2,3,5)\}$, $T_{10}=\{(1,5,6),(2,3,4)\}$. 
This is a perfect scheduling of a tournament with three teams in each match. For a general $k$, the following lemma provides a construction 
with  $O(k^2)$ rounds. 
\begin{lemma}
	There exists a partition $(T_1,\ldots,T_N)$ of $[k]\times[k]\times[k]$ for some $N\leq 24k^2$  such that
	$T_a$'s are disjoint subsets of  $[k]\times[k]\times[k]$,
	$\bigcup_{a\in[N]} T_a = [k]\times[k]\times[k]$,  $|T_a| \leq \lfloor k/3\rfloor  $ and
	for any $a\in[N]$ the set of random variables in $T_a$ satisfy
	\begin{eqnarray*}
		\{Y_{i,\ell,\ell',\ell''}\}_{i\in[d_1],(\ell,\ell',\ell'')\in T_a} \text{ are mutually independent }\;.
	\end{eqnarray*}
	\label{lmm:kwise_partition}
\end{lemma}
Now, we are ready to partition the summation. 
\begin{eqnarray}
	{\mathbb E}\big[ Z \big] &=& \frac{e^{-2\bb}}{2\,k^3\,d_1} {\mathbb E} \Big[  \sup_{\Delta\in\calB(D)} \sum_{i\in[d_1]} \sum_{\ell,\ell',\ell''\in[k]}
		\big\{\E[Y_{i,\ell,\ell',\ell''}(\Delta) ] - Y_{i,\ell,\ell',\ell''}(\Delta)\big\}  \Big] \nonumber \\
		&=&  \frac{e^{-2\bb}}{2\,k^3\,d_1}  {\mathbb E} \Big[ \sup_{\Delta\in\calB(D)} \sum_{i\in[d_1]}  \sum_{a\in[N]} \sum_{(\ell,\ell',\ell'') \in T_a }  \big\{\E[Y_{i,\ell,\ell',\ell''}(\Delta) ] - Y_{i,\ell,\ell',\ell''}(\Delta)\big\} \Big] \nonumber\\
		&\leq& \frac{e^{-2\bb}}{2\,k^3\,d_1}  \sum_{a\in[N]}   {\mathbb E} \Big[ \sup_{\Delta\in\calB(D)} \sum_{i\in[d_1]} \sum_{(\ell,\ell',\ell'') \in T_a }  \big\{\E[Y_{i,\ell,\ell',\ell''}(\Delta) ] - Y_{i,\ell,\ell',\ell''}(\Delta)\big\} \Big] \nonumber\\
		&\leq& \frac{e^{-2\bb}}{k^3\,d_1}  \sum_{a\in[N]}   {\mathbb E} \Big[ \sup_{\Delta\in\calB(D)} \sum_{i\in[d_1]} \sum_{(\ell,\ell',\ell'') \in T_a }   \xi_{i,\ell,\ell',\ell''} Y_{i,\ell,\ell',\ell''}(\Delta) \Big] \nonumber\\
		&=& \frac{e^{-2\bb}}{k^3\,d_1}  \sum_{a\in[N]}  {\mathbb E} \Big[ \sup_{\Delta\in\calB(D)} \sum_{i\in[d_1]} \sum_{(\ell,\ell',\ell'') \in T_a }   \xi_{i,\ell,\ell',\ell''} (\Delta_{i,j_{i,\ell}} - \Delta_{i,j_{i,\ell'}})^2 \chi_{i,\ell,\ell',\ell''}  \Big]\;,
		\label{eq:kwise_symbound3}
\end{eqnarray}
where the first inequality follows from the fact that sum of the supremum if no less than the supremum of the sum,
and the second inequality follows from standard symmetrization argument applied to independent random variables $\{Y_{i,\ell,\ell',\ell''}(\Delta) \}_{i\in[d_1],(\ell,\ell',\ell'')\in T_a}$  with i.i.d.\ Rademacher random variables $\xi_{i,\ell,\ell',\ell''}$'s.
Since $ (\Delta_{i,j_{i,\ell}} - \Delta_{i,j_{i,\ell'}})^2 \chi_{i,\ell,\ell',\ell''} \leq 4\bb  |\Delta_{i,j_{i,\ell}} - \Delta_{i,j_{i,\ell'}}| \chi_{i,\ell,\ell',\ell''}$,
we have by the Ledoux-Talagrand contraction inequality that
\begin{align}
	&{\mathbb E} \Big[ \sup_{\Delta\in\calB(D)} \sum_{i\in[d_1]} \sum_{(\ell,\ell',\ell'') \in T_a }   \xi_{i,\ell,\ell',\ell''} (\Delta_{i,j_{i,\ell}} - \Delta_{i,j_{i,\ell'}})^2  \chi_{i,\ell,\ell',\ell''} \Big] \nonumber\\
	& \leq \; 8\bb {\mathbb E} \Big[ \sup_{\Delta\in\calB(D)} \sum_{i\in[d_1]} \sum_{(\ell,\ell',\ell'') \in T_a }   \xi_{i,\ell,\ell',\ell''} \, \chi_{i,\ell,\ell',\ell''}\, \llangle \Delta, e_i(e_{j_{i,\ell}} - e_{j_{i,\ell'}})^T \rrangle  \Big] \label{eq:kwise_symbound2}
\end{align}
Applying H\"older's inequality, we get that
\begin{align}
	&\Big|  \sum_{i\in[d_1]} \sum_{(\ell,\ell',\ell'') \in T_a }   \xi_{i,\ell,\ell',\ell''} \,\chi_{i,\ell,\ell',\ell''} \, \llangle \Delta, e_i(e_{j_{i,\ell}} - e_{j_{i,\ell'}})^T \rrangle     \Big| \nonumber \\
	& \leq \; \nucnorm{\Delta}  \lnorm{ \sum_{i\in[d_1]} \sum_{(\ell,\ell',\ell'') \in T_a }   \xi_{i,\ell,\ell',\ell''} \,  \chi_{i,\ell,\ell',\ell''} \, \big( e_i(e_{j_{i,\ell}} - e_{j_{i,\ell'}})^T \big)   }{2} \;. 	\label{eq:kwise_symbound}
\end{align}

We are left to prove that the expected value of the right-hand side of the above inequality is bounded by
$C \nucnorm{\Delta}  \sqrt{k d_1 \log d / \min\{d_1,d_2\}} $ for some numerical constant $C$. 
For $i\in[d_1]$ and $(\ell,\ell',\ell'')\in T_a$, let $W_{i,\ell,\ell',\ell''} =\xi_{i,\ell,\ell',\ell''} \, \chi_{i,\ell,\ell',\ell''}\, \big( e_i(e_{j_{i,\ell}} - e_{j_{i,\ell'}})^T \big) $ 
be independent zero-mean random matrices, such that
\begin{eqnarray*}
	\lnorm{W_{i,\ell,\ell',\ell''}}{2} = \lnorm{\xi_{i,\ell,\ell',\ell''} \, \chi_{i,\ell,\ell',\ell''}\, \big( e_i(e_{j_{i,\ell}} - e_{j_{i,\ell'}})^T \big) }{2} \leq \sqrt{2} \;,
\end{eqnarray*}
almost surely, and
\begin{eqnarray*}
	\E[W_{i,\ell,\ell',\ell''}W_{i,\ell,\ell',\ell''}^T] &=& \E[\big( e_i(e_{j_{i,\ell}} - e_{j_{i,\ell'}})^T (e_{j_{i,\ell}} - e_{j_{i,\ell'}}) e_i^T\big)  \chi_{i,\ell,\ell',\ell''} ] \\
	&=& 2  \expect{ \chi_{i,\ell,\ell',\ell''} } e_ie_i^T \\
	&\preceq& 2 e_ie_i^T\;,
\end{eqnarray*}
and
\begin{eqnarray*}
	\E[W_{i,\ell,\ell',\ell''}^T W_{i,\ell,\ell',\ell''}]&=& \E[\big(  (e_{j_{i,\ell}} - e_{j_{i,\ell'}}) e_i^Te_i(e_{j_{i,\ell}} - e_{j_{i,\ell'}})^T\big)  \chi_{i,\ell,\ell',\ell''} ] \\
	&\preceq& \E[ (e_{j_{i,\ell}} - e_{j_{i,\ell'}}) e_i^Te_i(e_{j_{i,\ell}} - e_{j_{i,\ell'}})^T ] \\
	&=& \frac{2}{d_2} {\mathbf I}_{d_2 \times d_2} - \frac{2}{d_2^2} {\mathbf 1} {\mathbf 1}^T
	\;.
\end{eqnarray*}
This gives
\begin{eqnarray*}
	\sigma^2 &=& \max\left\{ \lnorm{ \sum_{i\in[d_1]} \sum_{(\ell,\ell',\ell'')\in T_a} \E[W_{i,\ell,\ell',\ell''}W_{i,\ell,\ell',\ell''}^T]}{2} , \lnorm{ \sum_{i\in[d_1]} \sum_{(\ell,\ell',\ell'')\in T_a} \E[W_{i,\ell,\ell',\ell''}^TW_{i,\ell,\ell',\ell''}]}{2} \right\} \\
	&\leq& \max\left\{ 2 |T_a|\,,\, \frac{2d_1 |T_a| }{d_2} \right\}   \,=\, \frac{2d_1|T_a|}{\min\{d_1,d_2\}} \leq \frac{2d_1 k }{3 \min\{d_1,d_2\}} \;,
\end{eqnarray*}
since we have designed $T_a$'s such that $|T_a|\leq k/3$.
Applying matrix Bernstein inequality \cite{Jo11} yields the tail bound
\begin{eqnarray*}
	\prob{\lnorm{\sum_{i\in[d_1]} \sum_{(\ell,\ell',\ell'')\in T_a} W_{i,\ell,\ell',\ell''} }{2} \geq t } &\leq& (d_1+d_2) \exp\Big( \frac{-t^2/2}{\sigma^2 + \sqrt{2} t/3} \Big) \;.
\end{eqnarray*}
Choosing $t= \max \big\{\, \sqrt{32 k d_1 \log d /(3\min\{d_1,d_2\})} , (16\sqrt{2}/3) \log d \,\big\}$,
we obtain with probability at least  $1-2d^{-3}$,
\begin{eqnarray*}
	\lnorm{\sum_{i\in[d_1]} \sum_{(\ell,\ell',\ell'')\in T_a} W_{i,\ell,\ell',\ell''} }{2} &\leq&
		\max \left\{ \sqrt{\frac{32 k d_1 \log d }{3 \min\{d_1 , d_2\}}} \,,\, \frac{16 \sqrt{2} \log d}{3} \right\}\;.
\end{eqnarray*}
It follows from the fact
$\lnorm{\sum_{i\in[d_1]}\sum_{(\ell,\ell',\ell'')\in T_a} W_{i,\ell,\ell',\ell''}}{2} \leq
\sum_{i,(\ell,\ell',\ell'')} \lnorm{W_{i,\ell,\ell',\ell''}}{2} \leq \sqrt{2} d_1 k /3$ that
\begin{eqnarray*}
	\E\left[\lnorm{\sum_{i\in[d_1]}\sum_{(\ell,\ell',\ell'')\in T_a} W_{i,\ell,\ell',\ell''}}{2}\right] &\leq&
		\max\left\{ \sqrt{\frac{32 k d_1 \log d }{3 \min\{d_1 , d_2\}}},  \frac{16 \sqrt{2} \log d}{3} \right\} \,+\,  \frac{ 2 \sqrt{2} d_1 k}{3d^3}\\
		&\leq & 2\sqrt{\frac{32 k d_1 \log d }{3 \min\{d_1 , d_2\}}} \;,
\end{eqnarray*}
where the last inequality follows from the assumption that $(16\min\{d_1,d_2\} \log d)/(3d_1)  \leq k\leq d_1^2 \log d $.
Substituting this in the RHS of Eq. \eqref{eq:kwise_symbound},
and then together with Eqs. \eqref{eq:kwise_symbound2} and \eqref{eq:kwise_symbound3},
this gives the following desired bound:
\begin{eqnarray*}
	\E[Z] &\leq& \sum_{a\in[N]} \sup_{\Delta\in\calB(D)} \frac{16 \bb e^{-2\bb} }{k^3\,d_1  } \sqrt{\frac{32 k d_1 \log d }{3 \min\{d_1,d_2\}}} \nucnorm{\Delta}\\
		&\leq& \sum_{a\in[N]}   \frac{e^{-4\bb}\sqrt{2}}{16\sqrt{3} k^2\,d_1\,d_2}  \underbrace{\Big(2^{10} e^{2\bb} \bb d_2 \sqrt{\frac{d_1 \log d}{k \min\{d_1,d_2\} } } \Big)}_{=\mu} \nucnorm{\Delta} \\
		&\leq & \frac{9 e^{-4 \bb} D^2}{40 d_1d_2}\;,
\end{eqnarray*}
where the last inequality holds because $N\leq 4 k^2$ and  $ \mu \nucnorm{\Delta} \leq D^2$.

\subsection{Proof of Lemma \ref{lmm:kwise_partition}}
Recall that $Y_{i,\ell,\ell',\ell''}(\Delta) = (\Delta_{i,j_{i,\ell}} - \Delta_{i,j_{i,\ell'}})^2  \chi_{i,\ell,\ell',\ell''}$, as defined in \eqref{eq:defY}. 
From the random utility model (RUM) interpretation of the MNL model presented in Section~\ref{sec:intro}, 
it is not difficult to show that $Y_{i,\ell,\ell',\ell''}$ and $Y_{i,\tell,\tell',\tell''}$ are 
mutually independent if the two triples $(\ell,\ell',\ell'')$ and $(\tell,\tell',\tell'')$ do not overlap, i.e., no index is present in both triples.   

Now, borrowing the terminologies from round robin tournaments, 
we construct a schedule for a tournament with $k$ teams where each match involve three teams. 
Let $T_{a,b}$ denote a set of triples playing at the same round, indexed by two integers $a\in\{3,\ldots,2k-3\}$ and $b\in\{5,\ldots,2k-1\}$. 
Hence, there are total $N=(2k-5)^2$ rounds. 

Each round $(a,b)$ consists of disjoint triples and is defined as 
\begin{eqnarray*}
	T_{a,b} &\equiv& \big\{(\ell,\ell',\ell'') \in [k]\times[k]\times[k] \,|\, \ell<\ell'<\ell'', \ell+\ell'=a, \text{ and } \ell'+\ell''=b \big\}\;.
\end{eqnarray*}

We need to prove that $(a)$ there is no missing triple; and $(b)$ no team plays twice in a single round. 
First, for any ordered triple $(\ell,\ell',\ell'')$, there exists $a\in\{3,\ldots,2k-3\}$ and $b\in\{5,\ldots,2k-1\}$ such that 
$\ell+\ell'=a$ and $\ell'+\ell''=b$. This proves that all ordered triples are covered by the above construction. 
Next, given a pair $(a,b)$, no two triples in $T_{a,b}$ can share the same team. 
Suppose there exists two distinct ordered triples $(\ell,\ell',\ell'')$ and $(\tell,\tell',\tell'')$ both in $T_{a,b}$, and one of the triples are shared.  
Then, from the two equations $\ell+\ell'=\tell+\tell'=a$ and $\ell'+\ell''=\tell'+\tell''=b $, it follows that all three indices must be the same, which 
is a contradiction. 
This proves the desired claim for ordered triples. 

One caveat is that we wanted to cover the whole $[k]\times[k]\times[k]$, and not just the ordered triples. 
In the above construction, for example, a triple $(3,2,1)$ does not appear. 
This can be resolved by simply taking all $T_{a,b}$'s from the above construction, and 
make 6 copies of each round, and permuting all the triples in each copy according to 
the same permutation over $\{1,2,3\}$. This increases the total rounds to $N=6(2k-5)^2\leq 24k^2$. 
Note that $|T_{a,b}|\leq\lfloor k/3\rfloor$ since no item can be in more than one triple. 

\section{Proof of estimating approximate low-rank matrices in Corollary \ref{cor:kwise_appxlowrank}}
\label{sec:kwise_cor_proof}
We follow closely the proof of a similar corollary in \cite{NW11}. 
First fix a threshold $\tau>0$, and set $r=\max\{j|\sigma_j(\Theta^*)>\tau\}$. With this choice of $r$, we have 
\begin{eqnarray*}
	\sum_{j=r+1}^{\min\{d_1,d_2\}} \sigma_j(\Theta^*) \;=\; \tau \sum_{j=r+1}^{\min\{d_1,d_2\}} \frac{\sigma_j(\Theta^*)}{\tau} \;\leq\; \tau \sum_{j=r+1}^{\min\{d_1,d_2\}} \Big(\frac{\sigma_j(\Theta^*)}{\tau}\Big)^q \;\leq\; \tau^{1-q} \rho_q \;.
\end{eqnarray*}
Also, since $r\tau^q \leq \sum_{j=1}^r \sigma_j(\Theta^*)^q \leq \rho_q$, it follows that 
$\sqrt{r} \leq \sqrt{\rho_a} \tau^{-q/2}$. Using these bounds, Eq. \eqref{eq:kwise_ub} is now 
\begin{eqnarray*}
	\fnorm{\hTheta-\Theta}^2\;\leq\;  \underbrace{288\sqrt{2}c_0 e^{4\bb}d_1d_2\lambda_0}_{ = A} \,\big( \sqrt{\rho_q} \tau^{-q/2} \fnorm{\hTheta-\Theta} + \tau^{1-q}\rho_q \,\big) \;.
\end{eqnarray*}
With the choice of  $\tau = A$, it follows after some algebra that 
\begin{eqnarray*}
	\fnorm{\hTheta-\Theta} \;\leq\; 2 \sqrt{ \rho_q } A^{(2-q)/2} \;.
\end{eqnarray*}

\section{Proof of the information-theoretic lower bound in Theorem \ref{thm:kwise_lb}}
\label{sec:kwise_lb_proof}

The proof uses information-theoretic methods 
which reduces the estimation problem to a multiway hypothesis testing problem. 
to prove a lower bound on the expected error, it suffices to prove 
\begin{eqnarray}
	\sup_{\Theta^*\in\Omega_\bb } \prob{ \fnorm{\hTheta-\Theta^*}^2 \geq \frac{\delta^2}{4} } &\geq& \frac{1}{2}
	\;.\label{eq:lb_fano1}
\end{eqnarray}
To prove the above claim, we follow the standard recipe of constructing a packing in $\Omega_\bb$. Consider a family $\{\Theta^{(1)},\ldots,\Theta^{(M(\delta)}\}$ of 
$d_1\times d_2$ dimensional matrices contained in $\Omega_\bb$ satisfying 
$\fnorm{\Theta^{(\ell_1)}-\Theta^{(\ell_2)}}\geq\delta$ for all $\ell_1,\ell_2,\in[M(\delta)]$. 
We will use $M$ to refer to $M(\delta)$ for simplify the notation. 
 Suppose we draw an index $L\in[M(\delta)]$ uniformly at random, and 
 we are given direct observations 
 $\sigma_i$ as per MNL model with $\Theta^*=\Theta^{(L)}$ on a randomly chosen set of $k$ items $S_i$ 
 for each user $i\in[d_1]$. 
It follows from triangular inequality that 
\begin{eqnarray}
	\sup_{\Theta^*\in\Omega_\bb } \prob{ \fnorm{\hTheta-\Theta^*}^2 \geq \frac{\delta^2}{4} } &\geq& \prob{\hL \neq L}\;,
	\label{eq:lb_fano2}
\end{eqnarray}
where $\hL$ is the resulting best estimate of the multiway hypothesis testing on $L$. 
 The generalized Fano's inequality gives 
 \begin{eqnarray}
 	\prob{\hL\neq L | S(1),\ldots,S(d_1)} &\geq& 1-\frac{I(\hL;L) + \log 2}{\log M} \\
	&\geq& 1-\frac{ {M \choose 2}^{-1} \sum_{\ell_1,\ell_2\in[M]} D_{\rm KL}(\Theta^{(\ell_1)} \| \Theta^{(\ell_2)}) +\log 2}{\log M} \label{eq:kwise_fano}\;,
 \end{eqnarray}
where $D_{\rm KL}(\Theta^{(\ell_1)}\|\Theta^{(\ell_2)})$ denotes the Kullback-Leibler divergence between the distributions of the partial rankings 
$\prob{\sigma_1,\ldots,\sigma_{d_1}|\Theta^{(\ell_1)},S(1),\ldots,S(d_1)}$ and 
$\prob{\sigma_1,\ldots,\sigma_{d_1}|\Theta^{(\ell_2)},S(1),\ldots,S(d_1)}$. 
The second inequality follows from a standard technique, which we repeat here for completeness. 
Let $\Sigma=\{\sigma_1,\ldots,\sigma_{d_1}\}$ denote the observed outcome of comparisons. 
Since $L\text{--} \Theta^{(L)}\text{--}\Sigma\text{--}\hL$ form a Markov chain, the data processing inequality gives 
$I(\hL;L) \leq I(\Sigma;L)$. For simplicity, we drop the conditioning on the set of alternatives $\{S(1),\ldots,S(d_1)\}$, and 
and let $p(\cdot)$ denotes joint, marginal, and conditional distribution of respective random variables. 
It follows that 
\begin{eqnarray} 
	I(\Sigma;L) &=& \sum_{\ell\in[M],\Sigma}  p(\Sigma|\ell)\frac{1}{M} \log \frac{p(\ell,\Sigma)}{p(\ell)p(\Sigma)} \nonumber\\
		&=& \frac{1}{M} \sum_{\ell\in[M]} \sum_{\Sigma}  p(\Sigma|\ell) \log \frac{p(\Sigma|\ell)}{\frac{1}{M}\sum_{\ell'}p(\Sigma|\ell')}    \nonumber\\
		&\leq& \frac{1}{M^2} \sum_{\ell,\ell'\in[M]} \sum_{\Sigma}  p(\Sigma|\ell) \log \frac{p(\Sigma|\ell)}{p(\Sigma|\ell')} \nonumber\\
		&=& \frac{1}{M^2} \sum_{\ell,\ell'\in[M]} D_{\rm KL} (\Theta^{(\ell_1)} \| \Theta^{(\ell_2)} )\;,
\end{eqnarray}
where the first inequality follows from Jensen's inequality.
To compute the KL-divergence, recall that from the RUM interpretation of the MNL model (see Section \ref{sec:intro}),  
one can generate sample rankings $\Sigma$  by drawing random variables with exponential distributions with mean $e^{\Theta^*_{ij}}$'s. Precisely, let  
$X^{(\ell)} = [X^{(\ell)}_{ij}]_{i\in[d_1],j\in S_i} $ denote the set of random variables, where 
$X^{(\ell)}_{ij}$ is drawn from the exponential distribution with mean $e^{-\Theta^{(\ell)}_{ij}}$. 
The MNL ranking follows by ordering the alternatives in each $S_i$ according to this $\{X^{(\ell)}_{ij}\}_{j\in S_i}$ 
by ranking the smaller ones on the top. This forms a 
Markov chain $L\text{--}X^{(L)}\text{--}\Sigma$, and the 
standard data processing inequality gives 
\begin{eqnarray}
	D_{\rm KL}(\Theta^{(\ell_1)}\|\Theta^{(\ell_2)} ) &\leq& D_{\rm KL}( X^{(\ell_1)} \|X^{(\ell_2)} ) \\ 
		&=& \sum_{i\in[d_1]} \sum_{j\in S_i}   \Big\{ e^{\Theta^{(\ell_1)}_{ij}-\Theta^{(\ell_2)}_{ij}} - (\Theta^{(\ell_1)}_{ij}-\Theta^{(\ell_2)}_{ij}) -1 \Big\} \\
		&\leq& \frac{e^{2\bb}}{4\bb^2} \sum_{i\in[d_1]} \sum_{j\in S_i}    (\Theta^{(\ell_1)}_{ij}-\Theta^{(\ell_2)}_{ij})^2 \;,
\end{eqnarray}
where the last inequality follows from the fact that $e^x-x-1 \leq (e^{2\bb}/(4\bb^2))x^2$ for any $x\in[-2\bb,2\bb]$.
Taking expectation over the randomly chosen set of alternatives,  
\begin{eqnarray}
	\E_{S(1),\ldots,S(d_1)} [D_{\rm KL}(\Theta^{(\ell_1)}\|\Theta^{(\ell_2)})] &\leq&  
	\frac{e^{2\bb}\,k }{4\,\bb^2\, d_2}\fnorm{\Theta^{(\ell_1)}-\Theta^{(\ell_2)}}^2\;.
\end{eqnarray}
Combined with \eqref{eq:kwise_fano}, we get that 
\begin{eqnarray}
	\prob{\hL\neq L} &=& \E_{S(1),\ldots,S(d_1)}[\prob{\hL\neq L|S(1),\ldots,S(d_1)} ]\\
		&\geq & 1- \frac{ {M \choose 2}^{-1}  \sum_{\ell_1,\ell_2\in[M]}  (e^{2\bb} k /(4\bb^2 d_2))\fnorm{\Theta^{(\ell_1)} - \Theta^{(\ell_2)}}^2 + \log 2}{\log M}\;,
\end{eqnarray}
The remainder of the proof relies on the following probabilistic packing. 
\begin{lemma}
	\label{lem:packing} 
	Let $d_2 \geq d_1 \geq 607$ be positive integers. 
	Then for each $r\in\{1,\ldots,d_1 \}$, and for any positive 
	$\delta>0$ there exists a family of $d_1\times d_2$ dimensional matrices 
	$\{\Theta^{(1)},\ldots,\Theta^{(M(\delta))}\}$ with cardinality 
	$M(\delta) = \lfloor (1/4)\exp(r d_2 /576)\rfloor$ such that each matrix is rank $r$ and the following bounds hold:
	\begin{eqnarray}
		\fnorm{\Theta^{(\ell)}}&\leq& \delta \;, \text{ for all }\ell\in[M]\\
		\fnorm{\Theta^{(\ell_1)} - \Theta^{(\ell_2)}} &\geq& \delta \;, \text{ for all } 
		\ell_1, \ell_2\in[M] \\
		\Theta^{(\ell)} &\in& \Omega_{\tilde{\bb}} \;, \text{ for all } \ell\in [M]\;, 
	\end{eqnarray}
	with $\tilde{\bb}=   (8\delta/d_2)\sqrt{2\log d} $ for $d=(d_1+d_2)/2$.
\end{lemma}

Suppose $\delta\leq \alpha  d_2/(8\sqrt{2\log d}) $ such that 
the matrices in the packing set are entry-wise bounded by $\bb$,  then the above lemma 
implies that $\fnorm{\Theta^{(\ell_1)} - \Theta^{(\ell_2)}}^2 \leq4\delta^2$, which gives 
\begin{eqnarray*}
	\prob{ \hL\neq L } &\geq& 1- \frac{\frac{e^{2\bb} k \delta^2}{\bb^2 d_2} + \log 2}{\frac{rd}{576} - 2\log 2}  \;\geq\; \frac{1}{2}\;,
\end{eqnarray*}
where the last inequality holds for $\delta^2 \leq (\bb^2 d_2/(e^{2\bb} k))((rd/1152) -2\log 2)$.  
If we assume $rd \geq 3195$ for simplicity, 
this bound on $\delta$ can be simplified to $\delta\leq \bb e^{-\bb} \sqrt{r\,d_2\, d/(2304\, k)}$.
Together with \eqref{eq:lb_fano1} and \eqref{eq:lb_fano2}, this proves that for all $\delta\leq\min\{ \bb d_2  /(8\sqrt{2\log d}),   \bb e^{-\bb}  \sqrt{\,r\,d_2\, d/(2304\, k)} \}$, 
\begin{eqnarray*}
	\inf_{\hTheta} \sup_{\Theta^*\in\Omega_{\bb}} \E\Big[\, \fnorm{\hTheta-\Theta^*} \,\Big] &\geq& \frac{\delta}{4}\;.
\end{eqnarray*}
Choosing $\delta$ appropriately to maximize the right-hand side finishes the proof of the desired claim.

\subsection{Proof of Lemma \ref{lem:packing}}
Following the construction in \cite{NW11}, 
we use probabilistic method to prove the existence of the desired family. 
We will show that the following procedure succeeds in producing the desired family with probability at least half, which proves its  existence. 
Let $d=(d_1+d_2)/2$, and suppose $d_2\geq d_1$ without loss of generality. 
For the choice of  $M'=e^{r d_2 /576}$, and for each $\ell \in [M']$, generate a rank-$r$ matrix $\Theta^{(\ell)}\in\reals^{d_1\times d_2}$ as follows: 
\begin{eqnarray}
	\Theta^{(\ell)} &=& \frac{\delta}{\sqrt{r d_2}} U(V^{(\ell)})^T \Big( \id_{d_2 \times d_2} -\frac{1}{d_2}\ones\ones^T \Big)\;, 
	\label{eq:defpacking}
\end{eqnarray}
where 
$U\in\reals^{d_1\times r}$ is a random orthogonal basis such that $U^TU=\id_{r\times r}$ and $V^{(\ell)} \in\reals^{d_2\times r}$ is a random matrix with each entry $V^{(\ell)}_{ij} \in\{-1,+1\}$ chosen independently and uniformly at random. 

By construction, notice that $\fnorm{\Theta^{(\ell)}} = (\delta/\sqrt{rd_2})\fnorm{(V^{(\ell)})^T(\id-(1/d_2)\ones\ones^T)}  \leq \delta$, 
since $\fnorm{V^{(\ell)} }=\sqrt{rd_2}$ and $(\id-(1/d_2)\ones\ones^T)$ is a projection which can only decrease the norm. 

Now, consider $\fnorm{\Theta^{(\ell_1)}-\Theta^{(\ell_2)}}^2 = 
(\delta^2/(r d_2))\fnorm{(\id-(1/d_2)\ones\ones^T)(V^{(\ell_1)} - V^{(\ell_2)}) }^2 \equiv f(V^{(\ell_1)},V^{(\ell_2)})$ 
which is a function over $2 r d_2$ i.i.d. random Rademacher variables $V^{(\ell_1)}$ and $V^{(\ell_2)}$ which define $\Theta^{(\ell_1)}$ and $\Theta^{(\ell_2)}$ respectively. 
Since $f$ is Lipschitz in the following sense, 
we can apply McDiarmid's concentration inequality. 
For all $(V^{(\ell_1)},V^{(\ell_2)})$ and $(\tV^{(\ell_1)},\tV^{(\ell_2)})$ that differ in only one variable, say 
$\tV^{(\ell_1)} = V^{(\ell_1)} + 2e_{ij}$, for some standard basis matrix $e_{ij}$, we have  
\begin{align}
	& \big| f(V^{(\ell_1)},V^{(\ell_2)})-f(\tV^{(\ell_1)},\tV^{(\ell_2)}) \big| \;=\; \nonumber\\
	& \;\; \left| \; \frac{\delta^2}{r\, d_2} \fnorm{ (\id-\frac{1}{d_2}\ones\ones^T) (V^{(\ell_1)} - V^{(\ell_2)})  }^2 \, 
	- \, \frac{\delta^2}{r\, d_2} \fnorm{ (\id-\frac{1}{d_2}\ones\ones^T) (V^{(\ell_1)} - V^{(\ell_2)} + 2e_{ij}) }^2 \; \right| \\
	& \;\; =\; \left|\; \frac{\delta^2}{r\,d_2} \fnorm{2(\id-\frac{1}{d_2}\ones\ones^T)e_{ij}}^2 + \frac{\delta^2}{r\,d_2} \llangle (\id-\frac{1}{d_2}\ones\ones^T) (V^{(\ell_1)} - V^{(\ell_2)})  , 2e_{ij} \rrangle \; \right| \\
	& \;\; \leq \; \frac{4\,\delta^2}{r\,d_2 }  \;+\; \frac{\delta}{r\,d_2}   \lnorm{(\id-\frac{1}{d_2}\ones\ones^T)  (V^{(\ell_1)} - V^{(\ell_2)}) }{\infty} \, \lnorm{2e_{ij} }{1} \\	
	& \;\; \leq \; \frac{12\,\delta^2}{r\,d_2 } \;,
\end{align}
where we used the fact that $(\id-\frac{1}{d_2}\ones\ones^T)(V^{(\ell_1)} - V^{(\ell_2)})  $ is entry-wise bounded by four. 
The expectation $\E[f(V^{(\ell_1)},V^{(\ell_2)})]$ is 
\begin{eqnarray}
	\frac{\delta^2}{r\,d_2}\E\left[\fnorm{(\id-\frac{1}{d_2}\ones\ones^T)(V^{(\ell_1)} - V^{(\ell_2)} ) }^2 \right] &=& 
	\frac{2\delta^2}{r\,d_2} \E\left[\fnorm{(\id-\frac{1}{d_2}\ones\ones^T)V^{(\ell_1)}  }^2 \right]\\ 
	&=& \frac{2\delta^2}{r\,d_2}\E\left[\fnorm{V^{(\ell_1)}}^2\right] - \frac{2\delta^2}{r\,d_2^2} \E\Big[\|\ones^TV^{(\ell_1)}\|^2\Big] \\
	&=& \frac{2\,\delta^2\,(d_2-1)}{d_2}\;. 
\end{eqnarray}
Applying McDiarmid's inequality with bounded difference $12\delta^2/(r d_2)$, we get that 
\begin{eqnarray}
	\prob{\, f(V^{(\ell_1)},V^{(\ell_2)}) \leq 2\delta^2(1-1/d_2) -t \, } &\leq& \exp\Big\{ - \frac{ t^2\,r\,d_2}{144 \,\delta^4 } \Big\} \;,
\end{eqnarray}
Since there are less than $(M')^2$ pairs of $(\ell_1,\ell_2)$, setting $t=(1-2/d_2)\delta^2$ and applying the union bound gives 
\begin{eqnarray}
	\prob{ \min_{\ell_1,\ell_2\in[M']} \fnorm{\Theta^{(\ell_1)}-\Theta^{(\ell_2)}}^2 \geq \delta^2  } &\geq& 1- \exp\Big\{ - \frac{r\,d_2}{144}\Big(1-\frac{2}{d_2}\Big)^2 + 2\log M'\Big\} \geq \frac78\;, 
	\label{eq:mcdiarmid1}
\end{eqnarray}
where we used $M'=\exp\{rd_2/576\}$ and $d_2\geq607$. 

We are left to prove that $\Theta^{(\ell)}$'s are in 
$\Omega_{(8\delta/d_2)\sqrt{2\log d_2}}$ as defined in \eqref{eq:defkwiseomega}. 
Since we removed the mean such that $\Theta^{(\ell)} \ones=0$ by construction, 
we only need to show that the maximum entry is bounded by $(8\delta/d_2)\sqrt{2\log d_2}$. 
We first prove an upper bound in \eqref{eq:lb_maxbound} for a fixed $\ell \in[M']$, 
 and use this to show that there exists a large enough subset of matrices satisfying this bound. 
From \eqref{eq:defpacking}, consider  $(UV^T)_{ij} =  \llangle u_i, v_j\rrangle$, where 
$u_i\in\reals^{r}$ is the first $r$ entries of a random vector drawn uniformly from the $d_2$-dimensional sphere, and 
$v_j\in\reals^r$ is drawn uniformly at random from $\{-1,+1\}^r$ with $\|v_j\| = \sqrt{r}$. 
Using Levy's theorem for concentration on the sphere \cite{Led01}, we have 
\begin{eqnarray}
	\prob{ |\llangle u_i,v_j \rrangle| \geq t } &\leq& 2 \exp \Big\{ -\frac{d_2 \,t^2}{8\,r} \Big\}\;. 
\end{eqnarray}
Notice that by the definition \eqref{eq:defpacking}, $\max_{i,j}|\Theta^{(\ell)}_{ij}| \leq (2\delta/\sqrt{r d_2}) \max_{i,j} |\llangle u_i ,v_j\rrangle|$. Setting $t=\sqrt{(32 r/d_2) \log d_2}$ and taking the union bound over all $d_1d_2$ indices, we get 
\begin{eqnarray} 
	\prob{\, \max_{i,j} |\Theta^{(\ell)}_{ij}|  \leq \frac{2\delta\sqrt{32 \log d_2} }{d_2}\,} &\geq& 1-2d_1d_2 \exp \Big\{ -4 \log d_2 \Big\} \;\geq \;\frac12 \;,
	\label{eq:lb_maxbound}
\end{eqnarray}
for a fixed $\ell\in[M']$. 
Consider the event that there exists a subset $S\subset[M']$ of cardinality $M=(1/4)M'$ with the same bound on maximum entry, then 
from \eqref{eq:lb_maxbound} we get 
\begin{eqnarray}
	\prob{ \exists S\subset [M']\text{ such that } \lnorm{\Theta^{(\ell)}}{\infty}\leq \frac{2\delta\sqrt{32\log d_2}}{d_2} \text{ for all }\ell\in S} &\geq& \sum_{m=M}^{M'} {M' \choose m} \Big(\frac12\Big)^m\;,
	\label{eq:unioncardinality}
\end{eqnarray}
which is larger than half for our choice of $M < M'/2$. 


\section{Proof of Theorem \ref{thm:bundle_ub}} 
\label{sec:bundle_ub_proof}

We use similar notations and techniques as the proof of Theorem \ref{thm:kwise_ub} in Appendix \ref{sec:kwise_ub_proof}. 
From the definition of $\cL(\Theta)$ in Eq. \eqref{eq:defbundleL}, we have for the true parameter $\Theta^*$, the gradient evaluated at the true parameter is 
\begin{eqnarray}
	\nabla \cL(\Theta^*) &=& -\frac{1}{n}\sum_{i=1}^n (e_{u_i} e_{v_i}^T - p_i) \;, 
\end{eqnarray}
where $p_i$ denotes the conditional probability of the MNL choice for the $i$-th sample. Precisely, 
$p_i=\sum_{j_1\in S_i}\sum_{j_2\in T_i} p_{j_1,j_2|S_i,T_i} e_{j_1}e_{j_2}^T$ where 
$p_{j_1,j_2|S_i,T_i}$ is the probability that the pair of items $(j_1,j_2)$ is chosen 
 at the $i$-th sample such that $p_{j_1,j_2|S_i,T_i} \equiv \prob{(u_i,v_i)=(j_1,j_2)| S_i,T_i} = e^{\Theta^*_{j_1,j_2}}/(\sum_{j'_1\in S_i,j'_2\in T_i} e^{\Theta^*_{j_1',j_2'}})$, 
 where $(u_i,v_i)$ is the pair of items selected by the $i$-th user among the set of pairs of  alternatives $S_i\times T_i$. 
 The Hessian can be computed as 
 \begin{align}
 &	\frac{\partial^2 \cL(\Theta)}{\partial\Theta_{j_1,j_2}\,\partial\Theta_{j_1',j_2'}} \;=\; \frac{1}{n} \sum_{i=1}^n \ind\big( (j_1,j_2) \in S_i\times T_i \big) \frac{\partial p_{j_1,j_2|S_i,T_i}}{\partial \Theta_{j_1',j_2'}}\\
		& \; = \frac{1}{n} \sum_{i=1}^n  \ind\big( (j_1,j_2),(j_1',j_2') \in S_i\times T_i \big) \,\Big( p_{j_1,j_2|S_i,T_i} \ind((j_1,j_2)=(j_1',j_2')) -
			p_{j_1,j_2|S_i,T_i}p_{j_1',j_2'|S_i,T_i} \Big)\;, \label{eq:bundle_hess} 
 \end{align}
We use $\nabla^2\cL(\Theta) \in \reals^{d_1d_2\times d_1d_2}$ to denote this Hessian. 
Let $\Delta = \Theta^*-\hTheta$ where $\hTheta$ is an optimal solution to the convex optimization in \eqref{eq:bundleopt}. 
We introduce the following  key technical lemmas.

Lemma \ref{lmm:kwise_deltabound} Eq. \eqref{eq:kwise_deltabound}

The following lemma provides a bound on the gradient
using the concentration of measure for sum of independent random matrices \cite{Jo11}. 
\begin{lemma}\label{lmm:bundle_gradient2}
	For any positive constant $c \ge 1$ and $n \geq  (4(1+c)e^{2\bb} d_1d_2 \log d)/\max\{d_1,d_2\}$, with probability at least $1-2 d^{-c}$,
	\begin{eqnarray}
		\lnorm{\nabla \calL(\Theta^\ast)}{2} &\leq&
		\sqrt{\frac{ 4(1+c) e^{2\bb} \max\{d_1,d_2\} \, \log d} { d_1\,d_2\,n  }} 
		\;.
	\end{eqnarray}
\end{lemma}
Since we are typically interested in the regime where 
the number of samples is much smaller than the dimension $d_1\times d_2$ of the problem, 
the Hessian is typically not positive definite. However, when we restrict our attention to the vectorized $\Delta$ with 
relatively small nuclear norm, then we can prove restricted strong convexity, which gives the following bound. 
\begin{lemma}[{\bf Restricted Strong Convexity for bundled choice modeling}]
\label{lmm:bundle_hessian2}
Fix any $\Theta \in \Omega'_\bb$ and assume $ (\min \{d_1,d_2\}/\min\{k_1,k_2\})\log d \leq n \leq \min\{d^5 \log d, k_1 k_2 \max\{d_1^2,d_2^2\}   \log d\} $.
	Under the random sampling model of the alternatives $\{j_{i a}\}_{i\in[n],a\in[k_1]}$ from the first set of items $[d_1]$, 
	$\{j_{i b}\}_{i\in[n],b\in[k_1]}$ from the second set of items $[d_2]$ and
	the random outcome of the comparisons described in section \ref{sec:intro},
	with probability larger than $1-2d^{-2^{25}}$,
	\begin{eqnarray}
		{\rm Vec}(\Delta)^\top \,\nabla^2\calL(\Theta)\, {\rm Vec}(\Delta)  &\geq& \frac{e^{-2\bb}}{8\, d_1\,d_2} \fnorm{\Delta}^2\;,
		\label{eq:bundle_hessian2}
	\end{eqnarray}
	for all $\Delta$ in $\calA'$ where
	\begin{eqnarray}
		\calA' = \Big\{ \Delta \in \reals^{d_1\times d_2} \,\big|\, \lnorm{\Delta}{\infty} \leq 2\bb\,, \, \sum_{j_1\in[d_1],j_2\in[d_2]}\Delta_{j_1j_2} = 0\,\text{ and } \fnorm{\Delta}^2 \geq \tmu \nucnorm{\Delta}
		\Big\} \;. \label{eq:bundle_defA}
	\end{eqnarray}
	with 
	\begin{eqnarray}
		\tmu &\equiv& 2^{10}\, \bb\,d_1 d_2 \sqrt{\frac{ \log d}{n\,\min\{d_1,d_2\}\,\min\{k_1,k_2\} }} \;.
	\label{eq:bundle_defmu}
	\end{eqnarray}
\end{lemma}

Building on these lemmas, the proof of Theorem \ref{thm:bundle_ub} is divided into the following two cases.
In both cases, we will show that 
\begin{align}
	\fnorm{\Delta}^2  \;\leq\;  12 \, e^{2\bb}  c_1 \lambda_1 \,d_1d_2\,  \nucnorm{\Delta} \;, 
	\label{eq:bundle_errorfrobeniusbound}
\end{align}
 with high probability. Applying Lemma \ref{lmm:kwise_deltabound} 
proves the desired theorem. We are left to show Eq. \eqref{eq:bundle_errorfrobeniusbound} holds.

\bigskip
\noindent{\bf Case 1: Suppose $\fnorm{\Delta}^2 \geq  \tmu \,\nucnorm{\Delta} $.}
With $\Delta=\Theta^* - \hTheta$, the Taylor expansion yields
\begin{align}
\calL(\widehat{\Theta})=\calL(\Theta^\ast) -  {\llangle \nabla \calL(\Theta^\ast), \Delta \rrangle} + \frac{1}{2} {\rm Vec}(\Delta) \nabla^2\calL(\Theta) {\rm Vec}^\top (\Delta),
\end{align}
where $\Theta=a \widehat{\Theta} + (1-a) \Theta^\ast$ for some $a \in [0,1]$.
It follows from Lemma \ref{lmm:bundle_hessian2} that with probability at least $1-2d^{-2^{25}}$,
\begin{align*}
	 \calL(\widehat{\Theta}) -\calL(\Theta^\ast) 
	&\; \ge\; - \lnorm{\nabla\calL(\Theta^\ast)}{2} \nucnorm{\Delta}+ \frac{  e^{-2\bb}}{8 \,d_1\,d_2} \fnorm{\Delta}^2\;.
\end{align*}
From the definition of $\widehat{\Theta}$ as an optimal solution of the minimization, we have
\begin{align*}
\calL(\widehat{\Theta})-  \calL(\Theta^\ast)  \;\le \; \lambda \left(  \nucnorm{\Theta^\ast} - \nucnorm{\widehat{\Theta}} \right) \;\le\; \lambda \nucnorm{\Delta}\;.
\end{align*}
By the assumption, we choose $\lambda\geq 8 \lambda_1$. 
 In view of Lemma \ref{lmm:bundle_gradient2}, this implies that 
  $\lambda \geq 2\lnorm{\nabla\calL (\Theta^*)}{2} $ 
  with probability at least $1-2d^{-3}$. 
   It follows that with probability at least $1-2d^{-3}- 2d^{-2^{25}}$,
\begin{align*}
\frac{ e^{-2\bb}}{8 d_1 d_2 } \fnorm{\Delta}^2 \;  \leq \;  \big(\lambda + \lnorm{\nabla\calL(\Theta^*)}{2}\big)\,  \nucnorm{\Delta} \;\leq\; \frac{3 \lambda}{2}  \nucnorm{\Delta} \;.
\end{align*}
By our assumption on $\lambda \leq c_1 \lambda_1$, this proves the desired bound in Eq. \eqref{eq:bundle_errorfrobeniusbound}

\noindent{\bf Case 2:  Suppose $\fnorm{\Delta}^2 \leq  \tmu \,\nucnorm{\Delta} $.}  
By the definition of $\mu$ and the fact that $c_1 \ge 128/\sqrt{\min\{k_1,k_2\}}$, it follows that 
$\tmu \le 12  \, e^{2\bb}  c_1 \lambda_1   \,d_1d_2$, and we get the same bound as in Eq. \eqref{eq:bundle_errorfrobeniusbound}.

\subsection{Proof of Lemma \ref{lmm:bundle_gradient2}}

Define $X_i=-(e_{u_i}e_{v_i}^T-p_i)$ such that $\nabla\cL(\Theta^*) = (1/n)\sum_{i=1}^n X_i$, which is a sum of 
$n$ independent random matrices. Note that since $p_i$ is entry-wise bounded by $e^{2\bb}/(k_1k_2)$, 
\begin{eqnarray*}
	\lnorm{X_i}{2} &\leq& 1 + \frac{e^{2\bb}}{\sqrt{k_1 k_2}} \;,
\end{eqnarray*}
and 
\begin{eqnarray}
	\sum_{i=1}^n \E[X_iX_i^T]  &= &  \sum_{i=1}^n (\E[e_{u_i}e_{u_i}^T] - p_ip_i^T)\\
		&\preceq& \sum_{i=1}^n \E[e_{u_i}e_{u_i}^T]\\
		& \preceq&\frac{e^{2\bb}\,n}{d_1} \id_{d_1\times d_1}\;,
\end{eqnarray}
where the last inequality follows from the fact that for any given $S_i$, 
$u_i$ will be chosen with probability at most $e^{2\bb}/k_1$, 
if it is in the set $S_i$ which happens with probability $k_1/d_1$.
Therefore, 
\begin{eqnarray}
	\lnorm{\sum_{i=1}^n \E[X_iX_i^T]}{2}  &\leq &  \frac{e^{2\bb} \,n}{d_1} \;.
\end{eqnarray}
Similarly, 
\begin{eqnarray}
	\lnorm{\sum_{i=1}^n \E[X_i^TX_i]}{2}  &\leq &  \frac{e^{2\bb} \,n}{d_2} \;.
\end{eqnarray}
Applying matrix Bernstein inequality \cite{Jo11}, we get 
\begin{eqnarray}
	\prob{ \lnorm{\nabla \cL(\Theta^*)}{2} > t} \leq (d_1+d_2) \exp\Big\{ \frac{- n^2 t^2/2}{(e^{2\bb}n\max\{d_1,d_2\}/(d_1d_2)) \,+\, ( (1 + (e^{2\bb}/\sqrt{k_1k_2}))n t /3)} \Big\}\;, 
\end{eqnarray}
which gives the desired tail probability of $2d^{-c}$ for the choice of 
\begin{eqnarray*}
	t &=& \max \Big\{ \sqrt{\frac{4(1+c) e^{2\bb} \max\{d_1,d_2\} \log d}{d_1d_2n}} \,,\, \frac{4(1+c) (1+\frac{e^{2\bb}}{\sqrt{k_1k_2}})\log d}{3n} \Big\} \\
	&=&\sqrt{\frac{4(1+c) e^{2\bb} \max\{d_1,d_2\} \log d}{d_1d_2n}} \;,
\end{eqnarray*}
where the last equality follows from the assumption that 
$n\geq (4(1+c) e^{2\bb} d_1d_2 \log d )/\max\{d_1,d_2\}$.

\subsection{Proof of  Lemma \ref{lmm:bundle_hessian2}}

Thee quadratic form of the Hessian defined in \eqref{eq:bundle_hess} can be lower bounded by 
\begin{eqnarray}
	{\rm Vec}(\Delta)^T\, \nabla^2\cL(\Theta)\, {\rm Vec}(\Delta)  &\geq & \underbrace{\frac{e^{-2\bb}}{2 \,k_1^2\,k_2^2\,n}\sum_{i=1}^n  \sum_{j_1,j_1'\in S_i}\sum_{j_2,j_2'\in T_i} \big( \Delta_{j_1,j_2}-\Delta_{j_1',j_2'} \big)^2  }_{\equiv \tH(\Delta)}\;,
	\label{eq:bundle_defH} 
\end{eqnarray}
which follows from Remark \ref{rem:hess}. 
To lower bound $\tH(\Delta)$, we first compute the mean: 
\begin{eqnarray}
	\E[\tH(\Delta)] &=& \frac{e^{-2\bb}}{2 \,k_1^2\,k_2^2\,n}\sum_{i=1}^n \E\big[ \sum_{j_1,j_1'\in S_i}\sum_{j_2,j_2'\in T_i} \big( \Delta_{j_1,j_2}-\Delta_{j_1',j_2'} \big)^2 \big]\\
	&=& \frac{e^{-2\bb}}{ \,d_1\,d_2}\fnorm{\Delta}^2 \;,
\end{eqnarray}
where we used the fact that $\E[\sum_{j_1\in S_i,j_2\in T_i}\Delta_{j_1,j_2}]=(k_1k_2/(d_1d_2))\sum_{j_1'\in[d_1],j_2'\in[d_2]}\Delta_{j_1',j_2'}=0$ for $\Delta \in \Omega'_{2\bb}$ in \eqref{eq:defbundleomega}. 

We now prove that $\tH(\Delta)$ does not deviate from its mean too much. 
Suppose there exists a $\Delta \in \calA'$ defined in \eqref{eq:bundle_defA} such that 
Eq. \eqref{eq:bundle_hessian2} is violated, i.e. $ \tH(\Delta) < (e^{-2\bb}/(8 k_1k_2d_1d_2))\fnorm{\Delta}^2$. 
In this case, 
\begin{eqnarray}
	\E[\tH(\Delta)] - \tH(\Delta)  &\geq& \frac{7\,e^{-2\bb}}{8 d_1d_2} \fnorm{\Delta}^2 \;.
\end{eqnarray}
We will show that this happens with a small probability. 
We use the same peeling argument as in Appendix \ref{sec:kwise_ub_proof} with 
\begin{eqnarray*}
	\calS'_\ell = \Big\{ \Delta\in\reals^{d_1\times d_2} \,|\, \lnorm{\Delta}{\infty} \leq 2\bb, \beta^{\ell-1}\tmu \leq \fnorm{\Delta}\leq\beta^\ell \tmu, \sum_{j_1\in[d_1],j_2\in[d_2]}\Delta_{j_1,j_2}=0, \text{ and }\nucnorm{\Delta}\leq \beta^{2\ell} \tmu \Big\}\;,
\end{eqnarray*}
where $\beta=\sqrt{10/9}$ and for $\ell\in\{1,2,3,\ldots\}$, and $\tmu$ is defined in \eqref{eq:bundle_defmu}.
By the peeling argument, there exists an $\ell\in \Z_+$ such that $\Delta\in\calS'_\ell$ and 
\begin{eqnarray}
	\E[\tH(\Delta)] - \tH(\Delta)  &\geq& \frac{7\,e^{-2\bb}}{8 d_1d_2} \beta^{2\ell-2}(\tmu)^2  \;\geq\;  \frac{7\,\,e^{-2\bb}}{9\, d_1d_2} \beta^{2\ell}(\tmu)^2 \;.
\end{eqnarray}
Applying the union bound over $\ell\in\Z_+$, 
\begin{align}
	&\prob{\exists \Delta \in \calA' \;,\; \tH(\Delta) < \frac{ e^{-2\bb} } { 8\, d_1\, d_2 } \fnorm{\Delta}^2 }\; \leq\;
		\sum_{\ell=1}^\infty \prob{ \sup_{\Delta\in\calS'_\ell} \big(\; \E[ \tH(\Delta)] - \tH(\Delta) \;\big) > \frac{7\,e^{-2 \bb} }{ 9 d_1 d_2}(\beta^\ell \tmu )^2 } \nonumber\\
		&\hspace{3.6cm}\leq\;\; \sum_{\ell=1}^\infty \prob{ \sup_{\Delta\in\calB'(\beta^\ell\tmu)} \big(\; \E[ \tH(\Delta)] - \tH(\Delta) \;\big) > \frac{7e^{-2 \bb}  }{ 9   d_1 d_2}(\beta^\ell \tmu)^2 }
		\;, \label{eq:bundle_peeling3}
\end{align}
where we define the set $\calB'(D)$ such that $\calS'_\ell \subseteq \calB'(\beta^\ell \tmu)$: 
\begin{align}
	\calB'(D) = \big\{\, \Delta \in \reals^{d_1\times d_2} \,\big|\, \|\Delta\|_\infty \leq 2\bb, \fnorm{\Delta}\leq D, \sum_{j_1\in[d_1],j_2\in[d_2]} \Delta_{j_1j_2}=0,  \tmu \nucnorm{\Delta} \leq D^2 \,
		  \Big\} \;. \label{eq:bundle_defcB}
\end{align}
The following key lemma provides the upper bound on this probability.  

\begin{lemma}
For $ (\min\{d_1,d_2\}/\min\{k_1,k_2\}) \log d \leq n \leq d^5\log d$, 
\begin{eqnarray}
	\prob{\sup_{\Delta\in\calB'(D)} \Big(\; \E[ \tH(\Delta)] - \tH(\Delta) \;\Big) \geq \frac{e^{-2\bb} D^2}{2 d_1d_2} } &\leq& \exp\Big\{ - \frac{ n\,\min\{k_1^2,k_2^2\}\,k_1k_2\,D^4}{2^{10} \bb^4 d_1^2 d_2^2 } \Big\}\;.
	\label{eq:bundle_hessian3}
\end{eqnarray}
\label{lmm:bundle_hessian3}
\end{lemma}

Let $\eta= \exp\left(-\frac{nk_1k_2 \min\{k_1^2,k_2^2\} (\beta-1.002) (\tmu)^4}{2^{10} \bb^4d_1^2 d_2^2} \right)$. 
Applying the tail bound to \eqref{eq:bundle_peeling3}, we get 
\begin{eqnarray*}
	\prob{\exists \Delta \in \calA' \;,\; \tH(\Delta) < \frac{ e^{-2\bb} } { 8  \,d_1 d_2 } \fnorm{\Delta}^2 } 
	&\leq & \sum_{\ell=1}^\infty \exp \Big\{
	- \frac{ n\,k_1k_2\,\min\{k_1^2,k_2^2\}\,(\beta^\ell \tmu)^4}{2^{10} \bb^4 d_1^2 d_2^2
	}\Big\} \\
	& \overset{(a)}{\leq} &\sum_{\ell=1}^\infty \exp\Big\{-\frac{n k_1k_2\min\{k_1^2,k_2^2\}\ell (\beta-1.002) (\tmu)^4 }{2^{10} \bb^4 d_1^2 d_2^2}\Big\} \\
	&\leq & \frac{\eta}{1-\eta},
\end{eqnarray*}
where $(a)$ holds because $\beta^{x} \geq x \log\beta \ge x(\beta-1.002)$ for the choice of $\beta=\sqrt{10/9}$.
By the definition of $\tmu$,
\begin{align*}
\eta \; = \; \exp\Big\{ - \frac{  2^{30}\,k_1 k_2  \max\{ d_2^2, d_1^2\} (\log d)^2 (\beta-1.002) }{ n}  \Big\} \;  \le\;   \exp \{ -\,2^{25}\,\log d\} \;,
\end{align*}
where the last inequality follows from the assumption that 
$n \leq  k_1 k_2  \max\{d_1^2,d_2^2\} \log d$, 
and $\beta - 1.002\geq 2^{-5}$. 
Since for $d \ge 2$, $\exp\{-2^{25}\log d\} \leq 1/2$ and thus $\eta \le 1/2$, the lemma follows by assembling the last two
displayed inequalities.

\subsection{Proof of Lemma \ref{lmm:bundle_hessian3}}
Let $Z \equiv \sup_{\Delta\in\calB'(D)} \E[\tH(\Delta)]-\tH(\Delta)$ and consider the tail bound using McDiarmid's inequality. 
Note that $Z$ has a bounded difference of 
$(8\bb^2e^{-2\bb}\max\{k_1,k_2\} )/(k_1^2k_2^2n)$ when one of the $k_1k_2n$ independent random variables are 
changed, which gives 
\begin{eqnarray}
	\prob{Z-\E[Z] \geq t} &\leq& \exp \Big(- \frac{k_1^4 k_2^4 n^2 t^2 }{64\bb^4 e^{-4\bb} \max\{k_1^2,k_2^2\}k_1k_2n} \Big)\;. 
\end{eqnarray}
With the choice of $t=D^2/(4e^{2\bb} \, d_1d_2)$, this gives 
\begin{eqnarray}
	\prob{Z-\E[Z] \geq \frac{e^{-2\bb}}{4 d_1d_2}D^2} &\leq& \exp \Big(- \frac{k_1^3 k_2^3 n D^4 }{2^{10} \bb^4  d_1^2d_2^2\max\{k_1^2,k_2^2\}} \Big)\;. 
\end{eqnarray}
We first construct a partition of the space similar to Lemma \ref{lmm:kwise_partition}. 
Let 
\begin{eqnarray}
	\tk &\equiv&  \min\{k_1,k_2\}   \;. \label{eq:bundle_deftk}
\end{eqnarray}
\begin{lemma}
	There exists a partition $(\cT_1,\ldots,\cT_N)$ of $\{[k_1]\times[k_2]\}\times\{[k_1]\times[k_2]\}$ for some $N\leq 2k_1^2k_2^2/\tk$  such that
	$\cT_\ell$'s are disjoint subsets,
	$\bigcup_{\ell\in[N]} \cT_\ell =  \{[k_1]\times[k_2]\}\times\{[k_1]\times[k_2]\}$,  $ |\cT_\ell| \leq \tk $ and
	for any $\ell \in[N]$ the set of random variables in $\cT_\ell$ satisfy
	\begin{eqnarray*}
		\{ (\Delta_{j_{i,a},j_{i,b}} - \Delta_{j_{i,a'},j_{i,b'}} )^2\}_{i\in[n],((a,b),(a',b'))\in \cT_\ell} \text{ are mutually independent }\;.
	\end{eqnarray*}
	where $j_{i,a}$ for $i\in[n]$ and $a\in[k_1]$ denote the $a$-th chosen item to be included in the set $S_i$. 
	\label{lmm:bundle_partition}
\end{lemma}

Now we prove an upper bound on $\E[Z]$ using the  symmetrization technique. 
Recall that $j_{i,a}$ is independently and uniformly chosen from $[d_1]$ for $i\in[n]$ and $a\in[k_1]$. 
Similarly, $j_{i,b}$ is independently and uniformly chosen from $[d_1]$ for $i\in[n]$ and $b\in[k_2]$. 
\begin{eqnarray}
	\E[Z] &=& 
	 \frac{e^{-2\bb}}{2 \,k_1^2\,k_2^2\,n} \E\left[ \sup_{\Delta\in\calB'(D)} \sum_{i=1}^n  \sum_{a,a'\in[k_1]}\sum_{b,b'\in[k_2]} \E\big[\big( \Delta_{j_{i,a},j_{i,b}}-\Delta_{j_{i,a'},j_{i,b'}} \big)^2 \big] - \big( \Delta_{j_{i,a},j_{i,b}}-\Delta_{j_{i,a'},j_{i,b'}} \big)^2  \right]\\
	 &\leq& \frac{e^{-2\bb}}{2 \,k_1^2\,k_2^2\,n} \sum_{\ell\in[N]} \E\left[ \sup_{\Delta\in\calB'(D)} \sum_{i=1}^n  \sum_{(j_1,j_2,j_1', j_2')\in \cT_\ell} \E\big[\big( \Delta_{j_1,j_2}-\Delta_{j_1',j_2'} \big)^2 \big] - \big( \Delta_{j_1,j_2}-\Delta_{j_1',j_2'} \big)^2  \right]\\
	 &\leq& \frac{e^{-2\bb}}{ \,k_1^2\,k_2^2\,n} \sum_{\ell\in[N]} \E\left[ \sup_{\Delta\in\calB'(D)} \sum_{i=1}^n  \sum_{(j_1,j_2,j_1', j_2')\in \cT_\ell} \xi_{i,j_1,j_2,j_1',j_2'} \big( \Delta_{j_1,j_2}-\Delta_{j_1',j_2'} \big)^2  \right]\;, \label{eq:bundle_conc1}
\end{eqnarray}
where the first inequality follows for the fact that the supremum of the sum is smaller than the sum of supremum, and the second inequality follows from standard symmetrization with i.i.d. Rademacher random variables $\xi_{i,j_1,j_2,j_1',j_2'}$'s. 
It follows from Ledoux-Talagrand contraction inequality that 
\begin{align}
	& \E\left[ \sup_{\Delta\in\calB'(D)} \sum_{i=1}^n  \sum_{(j_1,j_2,j_1', j_2')\in \cT_\ell} \xi_{i,j_1,j_2,j_1',j_2'} \big( \Delta_{j_1,j_2}-\Delta_{j_1',j_2'} \big)^2  \right] \\
	& \;\;\; \leq\; 8\bb\,
		\E\left[ \sup_{\Delta\in\calB'(D)} \sum_{i=1}^n  \sum_{(j_1,j_2,j_1', j_2')\in \cT_\ell} \xi_{i,j_1,j_2,j_1',j_2'} \big( \Delta_{j_1,j_2}-\Delta_{j_1',j_2'} \big)  \right] \\
	& \;\;\; \leq \; 8\bb\,		\E\left[ \sup_{\Delta\in\calB'(D)} \nucnorm{\Delta} \lnorm{\sum_{i=1}^n  \sum_{(j_1,j_2,j_1', j_2')\in \cT_\ell} \xi_{i,j_1,j_2,j_1',j_2'} \big( e_{j_1,j_2}-e_{j_1',j_2'} \big)}{2}  \right] \\
	&\;\;\;\leq \; \frac{8 \bb D^2}{\tmu}\E\left[   \lnorm{\sum_{i=1}^n  \sum_{(j_1,j_2,j_1', j_2')\in \cT_\ell} \xi_{i,j_1,j_2,j_1',j_2'} \big( e_{j_1,j_2}-e_{j_1',j_2'} \big)}{2}  \right] \;, \label{eq:bundle_conc2}
\end{align}
where the second inequality follows for the H\"older's inequality and 
 the last inequality follows from $\tmu\nucnorm{\Delta}\leq D^2$ for all $\Delta \in \calB'(D)$.
To bound the expected spectral norm of the random matrix, 
we use matrix Bernstein's inequality. 
Note that 
$\lnorm{ \xi_{i,j_1,j_2,j_1',j_2'} c }{2}\leq\sqrt{2}$ almost surely,
$\E[( e_{j_1,j_2}-e_{j_1',j_2'})( e_{j_1,j_2}-e_{j_1',j_2'})^T] \preceq (2/d_1)\id_{d_1 \times d_1}$, and 
$\E[( e_{j_1,j_2}-e_{j_1',j_2'})^T( e_{j_1,j_2}-e_{j_1',j_2'})] \preceq (2/d_2)\id_{d_2 \times d_2}$. 
It follows that $\sigma^2=2n|\cT_\ell|/\min\{d_1,d_2\} $, where $|\cT_\ell|\leq \min\{k_1,k_2\}$. 
It follows that 
\begin{eqnarray*}
	\prob{\lnorm{\sum_{i=1}^n  \sum_{(j_1,j_2,j_1', j_2')\in \cT_\ell} \xi_{i,j_1,j_2,j_1',j_2'} \big( e_{j_1,j_2}-e_{j_1',j_2'} \big)}{2}> t} &\leq& (d_1+d_2)\exp\Big\{\frac{-t^2/2}{\frac{2n\min\{k_1,k_2\}}{\min\{d_1,d_2\}} +\frac{\sqrt{2} t}{3}}\Big\}\;,
\end{eqnarray*}
Choosing $t=\max\{ \sqrt{64 n (\min\{k_1,k_2\}/\min\{d_1,d_2\}) \log d} ,(16\sqrt{2}/3)\log d\}$, we obtain a bound on the spectral norm of $t$ with probability 
at least $1-2d^{-7}$. 
From the fact that $\lnorm{\sum_{i=1}^n  \sum_{(j_1,j_2,j_1', j_2')\in \cT_\ell} \xi_{i,j_1,j_2,j_1',j_2'} \big( e_{j_1,j_2}-e_{j_1',j_2'} \big)}{2} \leq (n /\sqrt{2})\min\{k_1,k_2\}$, it follows that 
\begin{align}
&	\E\left[   \lnorm{\sum_{i=1}^n  \sum_{(j_1,j_2,j_1', j_2')\in \cT_\ell} \xi_{i,j_1,j_2,j_1',j_2'} \big( e_{j_1,j_2}-e_{j_1',j_2'} \big)}{2}  \right] \\
	&\;\;\; \leq\;  \max\Big\{ \sqrt{ \frac{64\,n\,\min\{k_1,k_2\}\log d}{\min\{d_1,d_2\}} } ,(16\sqrt{2}/3)\log d\Big\} + \frac{2n \min\{k_1,k_2\}}{\sqrt{2} d^7}\\
	& \;\;\; \leq \; \sqrt{ \frac{66\,n\,\min\{k_1,k_2\}\log d}{\min\{d_1,d_2\}}} 
\end{align}
which follows form the assumption that 
$n \min\{k_1,k_2\} \geq \min\{d_1,d_2\} \log d$ and 
$n\leq d^5\log d$.
Substituting this bound in \eqref{eq:bundle_conc1}, and \eqref{eq:bundle_conc2}, we get that 
\begin{eqnarray}
	\E[Z] &\leq& \frac{16 e^{-2\bb} \bb D^2}{\tmu }\sqrt{\frac{66 \log d}{n \min\{k_1,k_2\} \min\{d_1,d_2\}}}\\
		&\leq& \frac{e^{-2\bb} D^2 }{4\, d_1d_2 }\;. 
\end{eqnarray}

\section{Proof of the information-theoretic lower bound in Theorem \ref{thm:bundle_lb}}
\label{sec:bundle_lb_proof}
This proof follow closely the proof of Theorem \ref{thm:kwise_lb} in Appendix \ref{sec:kwise_lb_proof}. 
We apply the generalized Fano's inequality in the same way to get Eq. \eqref{eq:kwise_fano} 
\begin{eqnarray}
 	\prob{\hL\neq L} 
	&\geq& 1-\frac{ {M \choose 2}^{-1} \sum_{\ell_1,\ell_2\in[M]} D_{\rm KL}(\Theta^{(\ell_1)} \| \Theta^{(\ell_2)}) +\log 2}{\log M} \label{eq:bundle_fano}\;,
\end{eqnarray}

The main challenge in this  case is that we can no longer directly apply the RUM interpretation to compete 
$D_{\rm KL}(\Theta^{(\ell_1)} \| \Theta^{(\ell_2)})$. 
This will result in over estimating the KL-divergence,  
because this approach does not take into account that we  only take the top winner, out of those $k_1k_2$ alternatives. 
Instead, we compute the divergence directly, and provide an appropriate bound. Let the set of $k_1$ rows and $k_2$ columns chosen in one of the $n$ sampling be $S \subset [d_1]$ and $T \subset [d_2]$ respectively. Then,
\begin{eqnarray}
	D_{\rm KL}(\Theta^{(\ell_1)} \| \Theta^{(\ell_2)}) &\overset{(a)}{=}& \frac{n}{{d_1 \choose k_1}{d_2 \choose k_2}} \sum_{S, T} \sum_{\substack{i \in S \\ j \in T}} \frac{e^{\Theta_{ij}^{(\ell_1)}}}{\sum_{\substack{i' \in S \\ j' \in T}} e^{\Theta_{i'j'}^{(\ell_1)}}} \log\left(\frac{e^{\Theta_{ij}^{(\ell_1)}}\sum_{\substack{i' \in S \\ j' \in T}} e^{\Theta_{i'j'}^{(\ell_2)}}}{e^{\Theta_{ij}^{(\ell_2)}}\sum_{\substack{i' \in S \\ j' \in T}} e^{\Theta_{i'j'}^{(\ell_1)}}}\right)\\ 
		&\overset{(b)}{\leq}& \frac{n}{{d_1 \choose k_1}{d_2 \choose k_2}} \sum_{S, T} \left(\sum_{\substack{i , j }}\frac{{e^{2\Theta_{ij}^{(\ell_1)}}}\sum_{\substack{i', j' }}e^{\Theta_{i'j'}^{(\ell_2)}} - e^{\Theta_{ij}^{(\ell_1)} + \Theta_{ij}^{(\ell_2)}}\sum_{\substack{i' , j' }}e^{\Theta_{i'j'}^{(\ell_1)}}}{e^{\Theta_{ij}^{(\ell_2)}} \left(\sum_{\substack{i',j' }} e^{\Theta_{i'j'}^{(\ell_1)}} \right)^2  }\right) \\
		&\overset{(c)}{\leq}& \frac{ne^{2\alpha}}{k_1^2 k_2^2 {d_1 \choose k_1}{d_2 \choose k_2}} \sum_{S,T} \sum_{i,j} \left (e^{2\Theta_{ij}^{(\ell_1)}-\Theta_{ij}^{(\ell_2)} }  \sum_{i', j' }e^{\Theta_{i'j'}^{(\ell_2)}} - e^{\Theta_{ij}^{(\ell_1)}} \sum_{i', j' }e^{\Theta_{i'j'}^{(\ell_1)}} \right)
		\\
		& = & \frac{ne^{2\alpha}}{k_1^2 k_2^2 {d_1 \choose k_1}{d_2 \choose k_2}} \sum_{S, T} \left( \sum_{\substack{i' , j' }} e^{\Theta_{i'j'}^{(\ell_2)}} \sum_{\substack{i,j }} \frac{\left(e^{\Theta_{ij}^{(\ell_1)}} - e^{\Theta_{ij}^{(\ell_2)}}\right)^2}{e^{\Theta_{ij}^{(\ell_2)}}}  - \Big(\sum_{i,j}  ( e^{\Theta_{ij}^{(\ell_1)}} - e^{\Theta_{ij}^{(\ell_2)}} )\Big)^2 \right) \\
		&\overset{(d)}{\leq}& \frac{ne^{4\alpha}}{k_1 k_2 {d_1 \choose k_1}{d_2 \choose k_2}} \sum_{S, T} \sum_{\substack{i , j }}\left(e^{\Theta_{ij}^{(\ell_1)}} - e^{\Theta_{ij}^{(\ell_2)}}\right)^2\\
		&\overset{(e)}{\leq}& \frac{ne^{5\alpha}}{k_1 k_2 {d_1 \choose k_1}{d_2 \choose k_2}} \sum_{S, T} \sum_{\substack{i , j }}\left(\Theta_{ij}^{(\ell_1)} - \Theta_{ij}^{(\ell_2)}\right)^2\\
		&\overset{(f)}{=}& \frac{ne^{5\alpha}}{d_1 d_2} \fnorm{\Theta_{ij}^{(\ell_1)} - \Theta_{ij}^{(\ell_2)}}^2\\
\end{eqnarray}

Here $(a)$ is by definition of KL-distance and the fact that $S$, $T$ are chosen uniformly from all possible such sets and $(b)$ is due to the fact that $\log(x) \leq x-1$ with $x=({e^{\Theta_{ij}^{(\ell_1)}}\sum_{\substack{i' \in S,j' \in T}} e^{\Theta_{i'j'}^{(\ell_2)}}})/({e^{\Theta_{ij}^{(\ell_2)}}\sum_{\substack{i' \in S, j' \in T}} e^{\Theta_{i'j'}^{(\ell_1)}}})$. The constants at $(c)$   is due to the fact that each element of $\Theta^{(\ell_1)}$ is upper bounded by $\alpha$ and lower bounded by $-\alpha$. We can get $(d)$ by 
removing the second term which is always negative, and using the bond of $\alpha$.  $(e)$ is obtained because $e^x$ where $-\alpha \leq x \leq \alpha$ is Lipschitz continuous with Lipschitz constant $e^\alpha$. At last $(f)$ is obtained by simple counting of the occurrences of each $ij$. 
Thus we have,
\begin{eqnarray}
 	\prob{\hL\neq L} 
	&\geq& 1-\frac{ {M \choose 2}^{-1} \sum_{\ell_1,\ell_2\in[M]} \frac{ne^{5\alpha}}{d_1 d_2} \fnorm{\Theta_{ij}^{(\ell_2)} - \Theta_{ij}^{(\ell_2)}}^2 +\log 2}{\log M},
\end{eqnarray}

The remainder of the proof relies on the following probabilistic packing.

\begin{lemma}
	\label{lem:bundle_packing} 
	Let $d_2 \geq d_1$ be sufficiently large positive integers. 
	Then for each $r\in\{1,\ldots,d_1 \}$, and for any positive 
	$\delta>0$ there exists a family of $d_1\times d_2$ dimensional matrices 
	$\{\Theta^{(1)},\ldots,\Theta^{(M(\delta))}\}$ with cardinality 
	$M(\delta) = \lfloor (1/4)\exp(r d_2 /576)\rfloor$ such that each matrix is rank $r$ and the following bounds hold:
	\begin{eqnarray}
		\fnorm{\Theta^{(\ell)}}&\leq& \delta \;, \text{ for all }\ell\in[M]\\
		\fnorm{\Theta^{(\ell_1)} - \Theta^{(\ell_2)}} &\geq& \frac12\delta \;, \text{ for all } 
		\ell_1, \ell_2\in[M] \label{eq:packing_lower}\\
		\Theta^{(\ell)} &\in& \Omega'_{\tilde{\bb}} \;, \text{ for all } \ell\in [M]\;, 
	\end{eqnarray}
	with $\tilde{\bb}=   (8\delta/d_2)\sqrt{2\log d} $ for $d=(d_1+d_2)/2$.
\end{lemma}

Suppose $\delta\leq \alpha d_2/(8 \sqrt{2 \log d)}$ such that 
the matrices in the packing set are entry-wise bounded by $\bb$,  then the above lemma \ref{lem:bundle_packing} implies that $\fnorm{\Theta^{(\ell_1)} - \Theta^{(\ell_2)}}^2 \leq4\delta^2$, which gives 
\begin{eqnarray}
	\label{eq:prob-2item}
	\prob{ \hL\neq L } &\geq& 1- \frac{\frac{e^{5\bb}n4\delta^2}{d_1d_2} + \log 2}{\frac{rd_2}{576} - 2\log2}  \;\geq\; \frac{1}{2}\;,
\end{eqnarray}
where the last inequality holds for $\delta^2 \leq (r d_1 d_2^2/(1152e^{5\bb}n))$ and assuming $rd_2 \geq 1600$. Together with \eqref{eq:prob-2item} and \eqref{eq:packing_lower}, this proves that for all $\delta\leq\min\{ \alpha d_2/(8 \sqrt{2 \log d)}, r d_1 d_2^2/(1152e^{5\bb}n)\}$, 
\begin{eqnarray*}
	\inf_{\hTheta} \sup_{\Theta^*\in\Omega_{\bb}} \E\Big[\, \fnorm{\hTheta-\Theta^*} \,\Big] &\geq& \delta/4\;.
\end{eqnarray*}
Choosing $\delta$ appropriately to maximize the right-hand side finishes the proof of the desired claim. 
Also by symmetry, we can apply the same argument to get similar bound with $d_1$ and $d_2$ interchanged.

\subsection{Proof of Lemma \ref{lem:bundle_packing}} 
\label{sec:bundle_lb_proof}

We show that the following procedure succeeds in producing the desired family with probability at least half, which proves its  existence. 
Let $d=(d_1+d_2)/2$, and suppose $d_2\geq d_1$ without loss of generality. 
For the choice of  $M'=e^{r d_2 /576}$, and for each $\ell \in [M']$, generate a rank-$r$ matrix $\Theta^{(\ell)}\in\reals^{d_1\times d_2}$ as follows: 
\begin{eqnarray}
	\Theta^{(\ell)} &=& \frac{\delta}{\sqrt{r d_2}} U(V^{(\ell)})^T \Big( \id_{d_2 \times d_2} -\frac{\ones^TU(V^{(\ell)})^T\ones}{d_1d_2}\ones\ones^T \Big)\;, 
	\label{eq:defpacking}
\end{eqnarray}
where 
$U\in\reals^{d_1\times r}$ is a random orthogonal basis such that $U^TU=\id_{r\times r}$ and $V^{(\ell)} \in\reals^{d_2\times r}$ 
is a random matrix with each entry $V^{(\ell)}_{ij} \in\{-1,+1\}$ chosen independently and uniformly at random. 
By construction, notice that $\fnorm{\Theta^{(\ell)}} \leq (\delta/\sqrt{rd_2})\fnorm{U(V^{(\ell)})^T}  = \delta$. 

Now, by triangular inequality, we have 
\begin{eqnarray*}
    \fnorm{\Theta^{(\ell_1)}-\Theta^{(\ell_2)}} &\geq& 
    \frac{\delta}{\sqrt{r d_2}} \fnorm{U(V^{(\ell_1)} - V^{(\ell_2)})^T } -  \frac{\delta\,|\ones^T U(V^{(\ell_1)} - V^{(\ell_2)})^T\ones|}{d_1d_2\sqrt{r d_2}}  \fnorm{ \ones\ones^T } \\
    &\geq& \frac{\delta}{\sqrt{r d_2}} \underbrace{\fnorm{V^{(\ell_1)} - V^{(\ell_2)} }}_{A} \,-\,  \frac{\delta}{\sqrt{r\, d_1\,d_2^2}}\big(\underbrace{|\ones^T U(V^{(\ell_1)})^T\ones|}_{B} +|\ones^T U (V^{(\ell_2)})^T \ones|\big)   \;.
\end{eqnarray*}
We will prove that the first term is bounded by $A\geq \sqrt{rd_2}$ with probability at least $7/8$ for all $M'$ matrices, 
and we will show that we can find $M$ matrices such that the second term is bounded by 
$B\leq 8 \sqrt{2 r d_2 \log (32 r) \log (32 d)}$ with probability at least $7/8$. 
Together, this proves that with probability at least $3/4$, there exists $M$ matrices such that 
\begin{eqnarray*}
    \fnorm{\Theta^{(\ell_1)}-\Theta^{(\ell_2)}} &\geq& \delta \Big(1-\sqrt\frac{2^7 \log (32r) \,\log (32d)}{ d_1 d_2}\Big)  \; \geq \; \frac{1}{2}\delta\;,
\end{eqnarray*}
for all $\ell_1$, $\ell_2\in[M]$ and for sufficiently large $d_1$ and $d_2$. 

Applying similar McDiarmid's inequality as Eq. \eqref{eq:mcdiarmid1} in Appendix \ref{sec:kwise_lb_proof}, it follows that 
$A^2 \geq r d_2 $ with probability at least $7/8$ for $M'= e^{r d_2/576}$ and a sufficiently large $d_2$.  

To prove a bound on $B$, we will show that for a given $\ell$, 
\begin{eqnarray}
    \prob{|\ones^TU(V^{(\ell)})^T\ones| \leq 8 \sqrt{2 r d_2 \log (32 r) \log (32 d)}} \;\geq\; \frac78 \;. \label{eq:bundle_diffbound1}
\end{eqnarray}
Then using the similar technique as in \eqref{eq:unioncardinality}, it follows that we can find $M=(1/4)M'$ matrices 
all satisfying this bound and also the bound on the max-entry in \eqref{eq:maxentrybound2}. 
We are left to prove \eqref{eq:bundle_diffbound1}. We apply a series of concentration inequalities. Let $H_1$ be the event that $\{|\llangle V^{(\ell)}_i,\ones \rrangle| \leq \sqrt{2 d_2 \log(32 r)} \text{ for all } i\in[r]\}$.
Then, applying the standard Hoeffding's inequality, we get that 
$\prob{ H_1 } \geq 15/16$, where $V^{(\ell)}_i$ is the $i$-th column of $V^{(\ell)}$. 
We next change the variables and represent $\ones^T U$ as 
$\sqrt{d_1} u^T \tilde{U}$, where $u$ is drawn uniformly at random from the unit sphere and $\tilde{U}$ is a $r$ dimensional subspace drawn uniformly  at random. 
By symmetry, $\sqrt{d_1} u^T \tilde{U}$ have the same distribution as $\ones^T U$. 
Let $H_2$ be the event that $\{ | \llangle \tilde{U}_i,(V^{(\ell)})^T \ones \rrangle | \leq \sqrt{16 r (d_2/d_1) \log(32r)\log(32 d) } \text{ for all } i \in [d_1]\}$, where $\tilde{U}_i$ is the $i$-th row of $\tilde{U}$. 
Then, applying Levy's theorem for concentration on the sphere \cite{Led01}, we have 
$\prob{H_2 | H_1} \geq 15/16$.
Finally, let $H_3$ be the event that 
$\{| \sqrt{d_1} \llangle u , \tilde{U} (V^{(\ell)})^T \rrangle \ones | \leq 8\sqrt{2 r d_2 \log(32 r)\log(32 d)} \}$. 
Then, again applying Levy's concentration, we get 
$\prob{H_3|H_1,H_2} \geq 15/16$. 
Collecting all three concentration inequalities, we get that with probability at least $13/16$, 
$|\ones^TU(V^{(\ell)})^T\ones| \leq 8\sqrt{2rd_2 \log(32r) \log (32d)}$, which proves Eq. \eqref{eq:bundle_diffbound1}.

We are left to prove that $\Theta^{(\ell)}$'s are in 
$\Omega_{(8\delta/d_2)\sqrt{2\log d_2}}$ as defined in \eqref{eq:defbundleomega}. 
 Similar to Eq. \eqref{eq:lb_maxbound}, applying Levy's concentration gives 
\begin{eqnarray}
	\prob{\, \max_{i,j} |\Theta^{(\ell)}_{ij}|  \leq \frac{2\delta\sqrt{32 \log d_2} }{d_2}\,} &\geq& 1-2 \exp \Big\{ -2 \log d_2 \Big\} \;\geq \;\frac12 \;,
	\label{eq:maxentrybound2}
\end{eqnarray}
for a fixed $\ell\in[M']$. 
 Then using the similar technique as in \eqref{eq:unioncardinality},  
it follows that  there exists $M=(1/4)M'$ matrices 
 all satisfying this bound and also the bound on $B$ in Eq. \eqref{eq:bundle_diffbound1}.

\end{document}